\documentclass[times, review, 10pt]{elsarticle}

\usepackage{amssymb}
\usepackage{amsmath}
\usepackage{setspace}        
\usepackage{booktabs}        
\usepackage{algorithm}       
\usepackage{algpseudocode}   
\usepackage{float}           
\usepackage{threeparttable}  
\usepackage{multirow}        
\usepackage{url}

\journal{Pattern Recognition}

\begin{document}

\begin{frontmatter}



\title{Addressing Spatial Indistinguishability in Spatiotemporal Prediction via Optimal Transport-Guided Masking}

\author[inst1]{Guangyu Wang}
\author[inst2]{Jiawei Tong\corref{cor1}}

\affiliation[inst1]{organization={Data Science and Artificial Intelligence},
            addressline={Dongbei University of Finance and Economics},
            city={Dalian, Liaoning},
            country={China}}

\affiliation[inst2]{organization={Institute of Systems, Molecular \& Integrative Biology},
            addressline={University of Liverpool},
            city={Liverpool},
            country={United Kingdom}}



\cortext[cor1]{Jiawei Tong (\url{Jiawei.Tong@liverpool.ac.uk}) is the corresponding author.}




\begin{abstract}
Spatiotemporal prediction aims to learn discriminative representations from correlated temporal signals over spatial structures for accurate future inference. A central challenge is \emph{spatial indistinguishability}: different nodes may share similar historical patterns yet evolve toward divergent futures, severely degrading forecasting performance in real-world sensor networks. Existing embedding-based and graph neural network (GNN)-based approaches can partially detect such ambiguous nodes but rely on historical similarity, struggling to capture \emph{future behavioral divergence}. We propose \textbf{STOT} (\textbf{S}patio\textbf{T}emporal \textbf{O}ptimal \textbf{T}ransport), a self-supervised framework that resolves spatiotemporal ambiguity via structured masking guided by optimal transport. Our key idea treats indistinguishability as a \emph{disambiguation} problem: future states are inferred by exploiting concurrent spatial correlations and their time-varying similarity. We design a similarity-aware metric for dynamic inter-node relationships and an optimal transport-based masking strategy to emphasize ambiguous positions during pre-training. A batch consistency constraint preserves semantic coherence, while a random-walk masking mechanism promotes structured context exploration. Experiments on six real-world datasets show that STOT performs competitively with state-of-the-art baselines on the evaluated benchmarks and improved interpretability through transport-plan visualizations.
\end{abstract}




\end{frontmatter}


\section{Introduction}

Spatiotemporal data prediction through sensor networks is a classic multivariate time series forecasting task~\citep{tkdetraffic}, with broad applications in earth system forecasting~\citep{earth}, traffic flow prediction~\citep{LV2025111499}, and energy prediction~\citep{jiang2017energy}. Prediction accuracy plays a crucial role in effective decision-making and minimizing socioeconomic impacts across these tasks. However, for traditional machine learning methods such as Support Vector Regression~\citep{awad2015support}, Random Forest~\citep{rigatti2017random}, and Gradient Boosting Decision Trees~\citep{ke2017lightgbm}, achieving high prediction accuracy remains challenging because of the complex spatiotemporal relationships and nonlinear patterns inherent in these systems.

Notably, deep learning advances in spatiotemporal prediction have revolutionized this field with increasingly effective solutions, especially after identifying spatial indistinguishability as the key insight for model design. In the early stages, although Recurrent Neural Network (RNN) and Convolutional Neural Network (CNN)-based methods showed promise in capturing temporal and spatial patterns~\citep{pal2021rnn}, their limitations in incorporating topological structures still hindered prediction performance. To address this limitation, Graph Neural Networks (GNNs) overcame topological limitations by capturing complex relationships between spatial nodes~\citep{li2018diffusion}, yet their architectural complexity with uncertain performance gains still posed significant challenges for spatiotemporal prediction.
A key breakthrough came with the specific illustration of spatial indistinguishability~\citep{deng2021st} -- nodes with similar historical patterns exhibiting divergent future trends -- which directly led to a paradigm shift in spatiotemporal prediction. This insight inspired new directions in model design, where embedding-based methods~\citep{shao2022spatial} tackled spatial indistinguishability more directly through simpler architectures by constructing spatially distinguishable embeddings, avoiding the performance uncertainties of complex GNN-based methods~\cite{ZOU2026112784}.

Currently, predicting spatially indistinguishable nodes poses a critical challenge for accurate spatiotemporal prediction, as these nodes exhibit similar historical patterns but divergent future trends. This challenge is particularly significant given that such nodes dominate over 50\% of sensor networks in real-world applications~\citep{shao2023exploring}. Although existing embedding-based methods~\citep{shao2022spatial} have attempted to improve performance by distinguishing node identities, this approach fails to capture the future behavioral characteristics of these nodes: while distinguishing identities may prevent interference with stable nodes, it does not directly contribute to predicting their future states. Such limitations persist in other approaches as well; GNN-based methods isolate these nodes through topological structures, yet still fail to capture their inherent predictive characteristics. We provide a more detailed discussion of their limitations in Section \ref{relate}.

In this paper, we propose a novel theoretical framework to address the fundamental challenge of predicting spatially indistinguishable nodes in spatiotemporal systems. Our key insight is that while these nodes may exhibit weak temporal autocorrelation, their future states can be effectively inferred through concurrent spatial correlations. This theoretical foundation is supported by empirical observations across various spatiotemporal domains, where local perturbations propagate through the system with spatially and temporally varying effects, creating predictable patterns of influence. Leveraging this theoretical framework, we introduce several methodological innovations. First, we develop a quantitative metric for dynamically evaluating spatial indistinguishability, providing a rigorous foundation for identifying and analyzing these challenging nodes. Second, inspired by recent advances in self-supervised learning, particularly masked autoencoders (MAE), we design a specialized pre-training paradigm that focuses on reconstructing the states of spatially indistinguishable nodes using information from their distinguishable counterparts. To enhance the effectiveness of this approach, we incorporate two complementary constraints: a batch consistency policy that maintains semantic coherence across training instances and a random walk mechanism that ensures comprehensive exploration of the representation space. These components are unified within an optimal transport framework, leading to our proposed method: \underline{\textbf{S}}patio\underline{\textbf{T}}emporal \underline{\textbf{O}}ptimal \underline{\textbf{T}}ransport (STOT).
The main contributions of this paper can be summarized as follows:
\begin{itemize}
    \item We present a systematic theoretical analysis of spatial indistinguishability in spatiotemporal prediction, revealing the critical importance of predicting indistinguishable nodes' future states.

    \item We introduce a novel timestep-level metric for quantifying spatial indistinguishability, enabling a more precise characterization of spatiotemporal dynamics.

    \item We develop an optimal transport constraint framework that ensures both semantic continuity and diversity through batch consistency and random walk strategies.

    \item We validate our theoretical framework through extensive experiments on six real-world datasets, demonstrating competitive or superior performance on the evaluated benchmarks and providing insights into spatiotemporal prediction mechanisms.
\end{itemize}

\section{Related Work}
\label{relate}
This section reviews the mainstream approaches for spatiotemporal prediction and discusses their limitations in accurately predicting future states of spatially indistinguishable nodes.

\subsection{Graph-based Methods}
Graph-based methods leverage the topological relationships among sensor nodes to represent spatial interactions in time series data, enabling the design of advanced spatiotemporal models. For instance, STGCN~\citep{yu2018spatio} integrates graph convolutional networks with gated temporal convolutions to capture comprehensive spatiotemporal features, while STNN~\citep{9142387} further refines this approach by incorporating spatial and temporal attention mechanisms to address region-based dependencies and external factors. Recently, a prevalent view has emerged that predefined graphs may introduce biases or prove inadequate in certain scenarios. In response, GTS~\citep{GTS} proposes learning latent graph structures jointly with spatiotemporal dynamics, and DFDGCN~\citep{li2023dynamicfrequencydomaingraph} combines static graphs with dynamically learned adjacency matrices. More recently, LvSTformer~\cite{linWhenMultiviewMeets2025}, a transformer-based approach, has been introduced to capture evolving multi-scale spatiotemporal patterns across sensors.
From a data-driven perspective, the superior modeling of spatiotemporal relationships by latent graph structures cannot be solely attributed to capturing time-varying interaction patterns. We argue that the key lies in addressing spatially indistinguishable nodes. In predefined graphs, these nodes are connected to others, and their inherently unpredictable future patterns can induce information confusion during message passing. Latent graphs, by contrast, effectively sever these problematic connections, which leads to consistently better performance, a view supported by \cite{shao2023exploring}. Clearly, GNN-based methods struggle to predict the future states of spatially indistinguishable nodes when they remain connected to other nodes~\cite{KUMAR2024110721}, and empirical evidence shows that disconnecting these nodes enhances overall predictive accuracy.

\subsection{Embedding-based Methods}
Embedding-based methods address spatial indistinguishability by designing spatially distinguishable models. Further examining the mechanisms of representative approaches, STID~\citep{shao2022spatial} augments nodes with identity information in both temporal and spatial dimensions, enabling node differentiation. STID provides a more direct solution that achieves similar results to models that learn latent graphs, disconnecting nodes that are spatially indistinguishable from each other. Building upon this, STAEformer~\citep{liu2023spatio} further integrates these identity-aware embeddings with transformer architectures to enhance performance. Another line of work focuses more on capturing location-specific temporal dynamics. ST-WA~\citep{cirstea2022towards} learns location-specific encodings to capture unique patterns across time series, while ST-Norm~\citep{deng2021st} achieves a similar effect through carefully designed normalization methods, eliminating the need for additional embeddings.

More recently, HiMNet~\citep{dong2024heterogeneity} learns heterogeneous meta-parameters from clustered spatial and temporal embeddings, which captures node-level behavioral differences without relying on explicit identity inputs. A parallel and more recent trend concerns generative modeling for spatiotemporal systems. For instance, Net-Ev$^{2}$~\citep{Wang_2026} simulates network event evolution with a generative framework, and Diffusion-TS~\citep{yuan2024diffusionts} provides an interpretable diffusion model for general time series generation.

Embedding-based methods inherently rely on node differentiation, similar to latent graph learning, as they disconnect spatially indistinguishable nodes from others. However, they fail to effectively predict the future states of these spatially indistinguishable nodes.

\subsection{Pre-trained Methods}
Pre-trained methods enable models to receive longer historical inputs, which has received increasing attention in spatiotemporal forecasting, as extended input sequences enhance a model's robustness to sudden events and improve its generalization to unseen data~\citep{STDMAE}. To obtain superior hidden representations, researchers have explored using pre-training techniques on time series data. One promising technique is the masked autoencoder (MAE), which masks out random input patches and reconstructs the original input based on the unmasked patches. This approach has been successfully applied in various domains, with masked pre-training yielding significant improvements in downstream tasks in computer vision~\citep{bao2021beit} and natural language processing~\citep{lan2019albert}.
Several studies have explored the application of MAE in spatiotemporal forecasting. For instance, STEP~\cite{STEP} employs MAE to reconstruct long historical time series, enhancing the model's prediction capabilities. However, Ti-MAE~\citep{li2023timaeselfsupervisedmaskedtime} noted the inconsistency between contrastive learning during pre-training and downstream prediction tasks and proposed using MAE as an auxiliary task to address this issue. Similarly, PITS~\citep{lee2024learningembedtimeseries} tackled this problem by combining MAE with comparative learning to avoid conflicts between pre-training and downstream tasks. Building upon these works, STD-MAE~\citep{STDMAE} considered the practical challenges of spatiotemporal forecasting and employed two MAEs in both temporal and spatial dimensions to reconstruct historical sequences, concatenating the spatiotemporal representations as inputs for the downstream predictor. Taking a different approach, SimMTM~\citep{dong2023simmtm} connects masked modeling with manifold learning to learn deeper representations.
However, these methods overlook the inherent nature of spatiotemporal forecasting and fail to strengthen training for spatial indistinguishability, diminishing pre-training performance and reducing efficiency. In contrast, our proposed method, STOT, introduces a carefully designed spatial indistinguishability-guided masking strategy during pre-training. By reconstructing indistinguishable nodes through distinguishable ones, STOT more effectively captures the complex interactions within spatiotemporal data.

\section{Problem Formulation}
\label{prob}
We formulate the spatiotemporal system as a graph $G = (V,E,A)$, where $V=\{v_{i}\}_{i=1,2,\ldots,N}$ represents the set of $N$ nodes in the system. The edge set $E = \{e_{ij}\}$ captures the connectivity between node pairs, with the adjacency matrix $A\in\{0,1\}^{N\times N}$ encoding these relationships such that $A_{ij}=1$ if nodes $v_i$ and $v_j$ are connected ($e_{ij}\in E$), and $A_{ij}=0$ otherwise. The system dynamics are characterized by a feature matrix $X\in\mathbb{R}^{N\times T_{long}\times C}$, where $T_{long}$ represents the temporal dimension of historical observations in the pre-training stage, which is much longer than the historical window $T$ used in end-to-end prediction and $C$ is the feature dimension. At each timestep $t$, the state of the system is captured by $x_t\in\mathbb{R}^N$.

\section{Methodology}
\label{meth}
\subsection{Overall Architecture of STOT}
\begin{figure}[ht]
    \centering
    \includegraphics[width=\columnwidth]{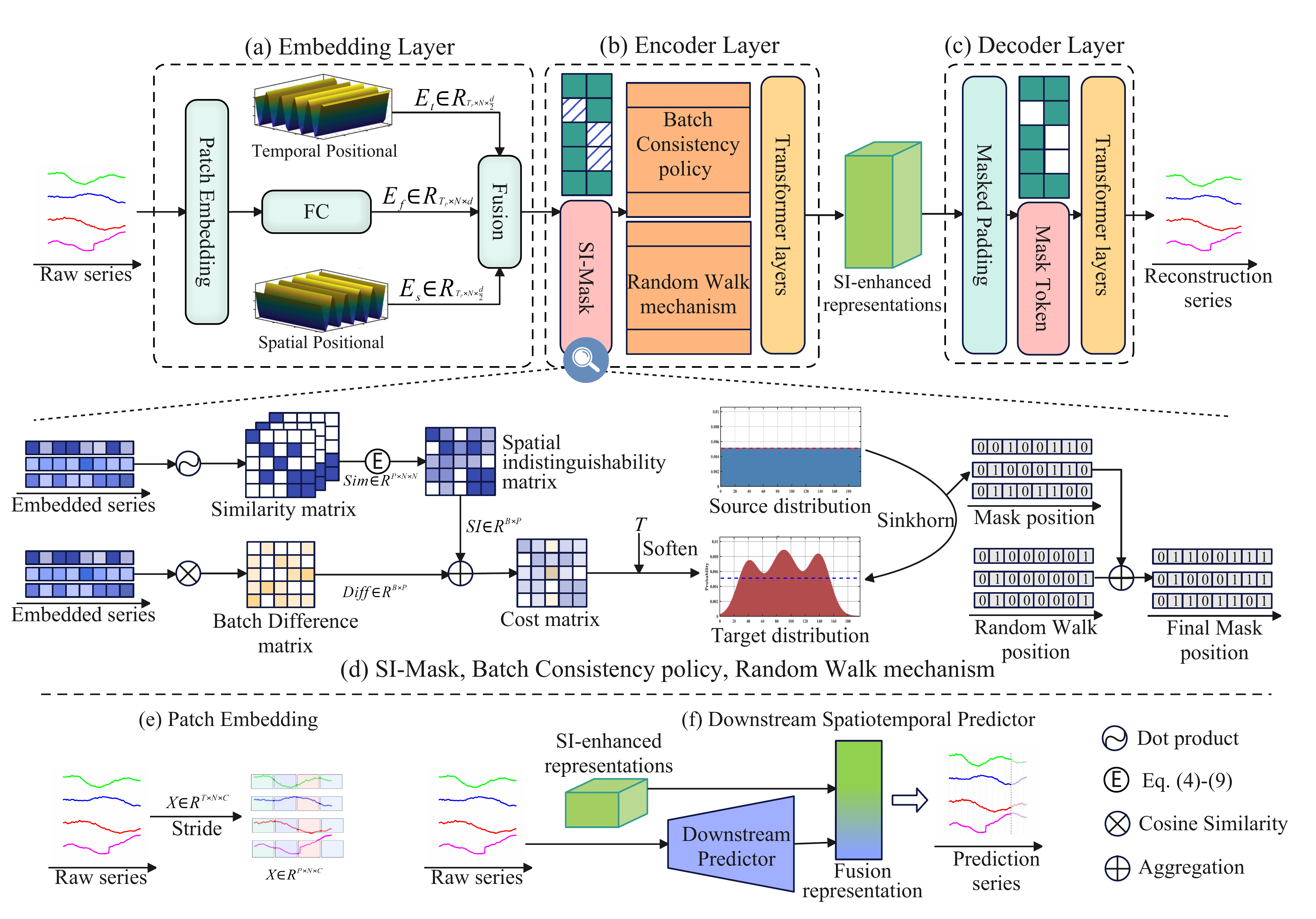}
    \caption{Architecture of STOT: (a) the embedding layer aggregates raw long-term traffic data with temporal and spatial positional features; (b) the encoder learns spatially indistinguishable node representations through masked reconstruction; (c) the decoder reconstructs masked positions from SI-enhanced representations; (d) mask generation uses SI-Mask, Batch Consistency, and Random Walk, where SI-Mask guides spatial indistinguishability and the latter two constrain representation continuity and richness; (e) patch embedding segments spatiotemporal data into local-pattern units; and (f) the STOT-enhanced downstream predictor can be integrated with various downstream architectures.}
    \label{fig:overall}
\end{figure}

As shown in Figure~\ref{fig:overall}, STOT contains two stages: pre-training (a)-(e) and downstream prediction (f). In pre-training, $T_{long}$ timesteps pass through an embedding layer that combines raw features with temporal and spatial positional encodings, and then through a Transformer encoder-decoder with an SI-Mask targeting spatially indistinguishable positions; the learned SI-enhanced representations are integrated with downstream predictors. Unlike random-mask MAE models, STOT needs only one pre-training phase without contrastive learning, achieving comparable representation quality more efficiently than STD-MAE~\citep{STDMAE} and PITS~\citep{lee2024learningembedtimeseries}.

\subsection{Embedding Layer}
The embedding layer produces position-aware representations through patch segmentation, high-dimensional feature projection, and temporal/spatial positional embeddings.

\subsubsection{Patch Segmentation}
Following~\cite{nie2023timeseriesworth64}, the input $X_{pre} \in \mathbb{R}^{T_{long} \times N \times C}$ is divided into $P$ non-overlapping patches of length $T_p = T_{long}/P$:
\begin{equation}
    X_{patch} = [\mathcal{P}_0(X_{pre}); \mathcal{P}_1(X_{pre}); ...; \mathcal{P}_{P-1}(X_{pre})],
\end{equation}
where $\mathcal{P}_i(\cdot)$ extracts the $i$-th segment, yielding $X_{patch} \in \mathbb{R}^{T_p \times N \times (CP)}$.

\subsubsection{Positional Embeddings}
Temporal Positional Embedding (TPE) encodes temporal position $p$ via sinusoidal functions:
\begin{equation}
    E_t(p,q,k) = \begin{cases}
        \sin(p/10000^{4k/D}), & \text{if } k \text{ is even} \\
        \cos(p/10000^{4k/D}), & \text{if } k \text{ is odd}
    \end{cases},
\end{equation}
yielding $E_t \in \mathbb{R}^{T_p \times N \times d/2}$. Spatial Positional Embedding (SPE) follows the same formulation with spatial position $q$ in place of $p$, yielding $E_s \in \mathbb{R}^{T_p \times N \times d/2}$.

\subsubsection{Embedding Output}
Raw data is projected via a fully connected layer $E_F = \text{FC}(X_{patch}) \in \mathbb{R}^{P \times N \times d}$. The final embedding concatenates all three components:
\begin{equation}
    Z = E_F \parallel E_t \parallel E_s \in \mathbb{R}^{T_p \times N \times 2d}.
\end{equation}

\subsection{Spatiotemporal Masked Pre-training}
Following the MAE architecture, the encoder masks spatially indistinguishable positions and the decoder reconstructs them; our masking strategy targets such positions, balances semantic richness with computational efficiency, and learns transferable spatiotemporal representations.

\subsubsection{Encoder Layer}
As shown in Figure~\ref{fig:overall}(b), the encoder masks positions with high spatial indistinguishability. Unlike prior dataset-level measurements~\cite{shao2023exploring}, our dynamic assessment captures time-varying node relationships and guides mask generation through an optimal transport (OT) formulation.

\textbf{Similarity Matrix Computation}
We compute cosine similarity matrices for both historical and future timesteps:
\begin{equation}
S^P_{t,i,j} = \frac{X^P_{t,i} \cdot X^P_{t,j}}{\|X^P_{t,i}\| \|X^P_{t,j}\|}, \quad S^F_{t,i,j} = \frac{X^F_{t,i} \cdot X^F_{t,j}}{\|X^F_{t,i}\| \|X^F_{t,j}\|}, \quad i,j \in \{1,\dots,N\}.
\end{equation}

\textbf{Spatial Indistinguishability Computation}
Given thresholds $\epsilon_u$ (high similarity) and $\epsilon_l$ (low similarity), the indistinguishability score for node $i$ at timestep $t$ is:
\begin{equation}
    I_{t,i} = \frac{\sum_{j \neq i} \mathbb{1}(S_{t,i,j} < \epsilon_l)}{\sum_{j \neq i} \mathbb{1}(S_{t,i,j} > \epsilon_u) + 1},
\end{equation}
which is high when node $i$'s similarity to others is uniformly distributed. We adopt this ratio between confidently dissimilar and confidently similar neighbors because it is directional, growing precisely when a node lacks reliable similar anchors yet is surrounded by dissimilar ones, which is the configuration that makes its future hard to infer. A dispersion summary such as the entropy of the similarity distribution is direction-agnostic and assigns the same value to an anchor-rich node and to an anchor-poor one, so it cannot separate easy nodes from genuinely ambiguous ones. The same construction also keeps the score insensitive to the exact thresholds, since it reflects the gap between the two groups rather than their absolute counts, and Figure~\ref{fig:formula5_kde} confirms that nodes with $I_{t,i}>0$ remain consistently harder to predict as $\epsilon_l$ varies. The overall score aggregates across both time periods and nodes:
\begin{equation}
    I_i = \alpha \cdot \frac{1}{T_p} \sum_{t=1}^{T_p} I^P_{t,i} + (1-\alpha) \cdot \frac{1}{T_{pred}} \sum_{t=1}^{T_{pred}} I^F_{t,i}, \quad
    Tscore = \frac{1}{N} \sum_{i=1}^N I_i,
\end{equation}
where $\alpha \in [0,1]$ balances historical and future contributions.

\textbf{Generating SI-Mask Position}
Directly selecting the top-$K$ highest $Tscore$ positions has two drawbacks: dispersed within-batch masks hinder coherent learning, and fixed masks for identical inputs limit semantic diversity. We address both by formulating mask generation as an OT problem.

Positions with high $Tscore$ are assigned lower transport cost via a log transformation, and a batch consistency penalty discourages overly dispersed masks:
\begin{equation}
    C_{base} = -\log(Tscore + 1), \quad C_{final} = C_{base} + \lambda \cdot \left|C_{base} - \frac{1}{B}\sum_{b=1}^B C_b\right|,
\end{equation}
where $\lambda$ controls the batch consistency weight.

The Sinkhorn algorithm~\citep{genevay2018learning} solves the OT problem by iteratively normalizing rows and columns to produce a doubly stochastic transport matrix (Pseudocode~\ref{pseudo}):

\begin{algorithm}[H]
\caption{Sinkhorn Algorithm for Mask Position Generation}
\begin{algorithmic}[1]
\Require Cost matrix $C_{final}$, number of iterations $K$, regularization parameter $\epsilon$
\Ensure Optimal transport plan $M$ (mask positions)
\State Initialize $M^{(0)} = \exp(-C_{final}/\epsilon)$
\For{$k = 1$ to $K$}
    \State $Q^{(k)} = M^{(k-1)}$ \Comment{Store previous iteration}
    \State $M^{(k)} = \text{Diag}(u^{(k)}) \cdot Q^{(k)} \cdot \text{Diag}(v^{(k)})$ \Comment{Update transport plan}
    \State $u^{(k)} = a \oslash (Q^{(k)}v^{(k-1)})$ \Comment{Row normalization}
    \State $v^{(k)} = b \oslash ((Q^{(k)})^T u^{(k)})$ \Comment{Column normalization}
\EndFor
\State \Return $M^{(K)}$ \Comment{Final transport plan as mask positions}
\end{algorithmic}
\label{pseudo}
\end{algorithm}
Mask positions are selected by choosing the $M_{total} = \lfloor T_p \cdot r_{mask} \rfloor$ entries that receive the largest transport mass per sample:
\begin{equation}
    \mathcal{M}_{b,s} = \begin{cases}
        1, & \text{if } s \in \text{topk}(M_b, M_{total}) \\
        0, & \text{otherwise}
    \end{cases},
\end{equation}
where $\text{topk}(\cdot, M_{total})$ returns the indices of the $M_{total}$ largest transport values for sample $b$. The cost is evaluated per sample and candidate timestep, so $C_{final}\in\mathbb{R}^{B\times T_p}$ with $C_{final,b,s}$ the cost of sample $b$ at timestep $s$, and $M_b$ is the $b$-th row of the transport plan $M^{(K)}$ returned by Algorithm~\ref{pseudo}, holding the mass that sample $b$ assigns to each candidate position. Since $M^{(0)}=\exp(-C_{final}/\epsilon)$ maps low cost to high mass, selecting the largest-mass entries of $M_b$ and setting $\mathcal{M}_{b,s}=1$ turns the transport plan into the binary SI-Mask, while the batch-consistency term and column normalization keep this selection coupled across samples.

\textbf{Random Walk Mechanism}
Controlled by $w_{ratio}$, random walk replaces $N_{replace}=\lfloor M_{total}\cdot w_{ratio}\rfloor$ positions in the OT-generated mask by sampling a new timestep $t_{new}$ from unmasked positions $\mathcal{T}\setminus\mathcal{M}_{batch}$ and releasing an existing mask position $p_{old}$; to maintain batch consistency, the operation is applied uniformly across samples:
\begin{equation}
    \mathcal{M}_{b,p_{old}} = 0, \quad \mathcal{M}_{b,t_{new}} = 1, \quad \forall b \in \{1,\ldots,B\},
\end{equation}
yielding the final binary mask $\mathcal{M} = [\mathcal{M}_{b,1}, \ldots, \mathcal{M}_{b,T_p}]$ that combines OT-guided positions with random walk diversity.

\textbf{Encoder with Mask} Given the masked input where $[MASK]$ tokens replace positions with $\mathcal{M}_{b,t}=1$:
\begin{equation}
    \tilde{x}_{b,t} = \begin{cases}
        [MASK], & \text{if } \mathcal{M}_{b,t} = 1 \\
        x_{b,t}, & \text{if } \mathcal{M}_{b,t} = 0
    \end{cases},
\end{equation}
the encoder applies standard Transformer layers—multi-head self-attention, feed-forward networks, and layer normalization—to produce the SI-enhanced representation after $L$ layers:
\begin{equation}
    \mathcal{R}^{SI} = \text{EncoderStack}_L(\tilde{X}_{patch}, \mathcal{M}) \in \mathbb{R}^{B \times T_p \times D}.
\end{equation}

\subsubsection{Decoder Layer}
The decoder uses the same Transformer architecture as the encoder; after padding to uniform dimensionality, $L_d$ decoder layers reconstruct masked positions from $\mathcal{R}^{SI}$ and yield $\hat{X}_{patch}\in\mathbb{R}^{B\times T_p\times D}$ with reconstructed values $Y_{pre}$ where $\mathcal{M}_{b,t}=1$.

\subsection{Downstream Spatiotemporal Prediction}
The pre-trained encoder processes $T_{long}$ timesteps to produce $\mathcal{R}^{SI}$, which is fused with the short-term representation from downstream predictor $F_0$ via an MLP:
\begin{equation}
    \mathcal{R}^{fused} = \text{MLP}([\mathcal{R}^{SI}; F_0(T)]).
\end{equation}
In this work we use GWNet~\citep{wu2019graph} as the primary predictor, with additional experiments on LSTM and ASTGCN~\citep{guo2019attention} to validate generalizability across architectures. Since this fusion runs after representation learning, where the pre-trained encoder already supplies informative spatial indistinguishability representations, we keep it as a simple MLP that preserves their integrity while remaining accurate, following classic models such as STAEformer~\citep{liu2023spatio}.

\subsection{Loss Function}
STOT uses two losses across its two training phases, with the encoder frozen during prediction:
\begin{equation}
    \mathcal{L}_{pre} = \frac{1}{B}\sum_{b=1}^B\sum_{t=1}^{T_p}\mathcal{M}_{b,t}||Y_{pre}^{b,t} - X_{patch}^{b,t}||_2^2, \quad
    \mathcal{L}_{pred} = \sqrt{\frac{1}{B}\sum_{b=1}^B||F_0(\mathcal{R}^{fused}) - Y||_2^2},
\end{equation}
where $\mathcal{L}_{pre}$ is the masked reconstruction loss over pre-training and $\mathcal{L}_{pred}$ is the RMSE for downstream forecasting.

\section{Experiments}
\label{exper}
We answer seven research questions: \textbf{(I)} STOT's performance against SOTA methods; \textbf{(II)} the effect of each mask pre-training component; \textbf{(III)} performance under high spatial indistinguishability; \textbf{(IV)} hyperparameter sensitivity across datasets; \textbf{(V)} computational efficiency versus other pre-training models; \textbf{(VI)} interpretability from OT visualization; and \textbf{(VII)} the effectiveness of the random walk mechanism compared with variants that change its unit or position.

\subsection{Experimental Settings}
\subsubsection{Datasets}
\begin{table}[ht]
\centering
    \caption{Dataset description and statistics.}
\renewcommand{\arraystretch}{1.2}
\setlength{\tabcolsep}{4pt}
\resizebox{\textwidth}{!}{
\begin{tabular}{lcccccl}
\hline
\textbf{Datasets} & \textbf{\#Sensors} & \textbf{\#Edges} & \textbf{\#Time Period} & \textbf{\#Type} & \textbf{\#Place} \\ \hline
PeMS03    & 358 & 574  & 2018/09--2018/11 & Flow  & North Central California, USA      \\
PeMS04    & 307 & 340  & 2018/01--2018/02 & Flow  & San Francisco Bay Area, USA         \\
PeMS07    & 883 & 866  & 2017/05--2017/08 & Flow  & Los Angeles area, USA          \\
PeMS08    & 170 & 295  & 2016/07--2016/08 & Flow  & San Bernardino area, USA   \\
PeMS-BAY  & 325 & 2369 & 2017/01--2017/06 & Speed & San Francisco Bay Area, USA \\
METR-LA   & 207 & 1515 & 2012/03--2012/06 & Speed & Los Angeles area, USA      \\ \hline
\end{tabular}}
\label{datasets}
\end{table}
We conducted extensive experiments using six real-world traffic datasets, which are classic and widely used examples of spatiotemporal data~\citep{afrin2020survey,cirstea2021enhancenet,choi2022STGNCDE}, with details provided in Table \ref{datasets}. The datasets are briefly described as follows: \textbf{PeMS03} provides traffic flow data from 358 sensors in California's District 3, spanning September to November 2018; \textbf{PeMS04} provides traffic flow data from 307 sensors in California's District 4, collected between January and February 2018; \textbf{PeMS07} provides traffic flow measurements from 883 sensors in California's District 7, covering May to August 2017; \textbf{PeMS08} provides traffic flow data from 170 sensors in California's District 8, obtained during July and August 2016; \textbf{METR-LA} provides traffic speed measurements from 207 sensors in Los Angeles, California, collected between March and June 2012; and \textbf{PeMS-BAY} provides traffic speed data from 325 sensors in the San Francisco Bay Area, California, spanning January to June 2017.
The traffic flow and speed datasets are partitioned into training, validation, and test subsets to facilitate model evaluation. For the traffic flow data, the ratio of the training, validation, and test subsets is 6:2:2, while for the traffic speed data, the ratio is 7:1:2.
\subsubsection{Baselines}


We compare STOT with 19 representative baselines covering classical statistical methods and recent SOTA deep learning models: ARIMA~\cite{ARIMA_pred} and VAR~\cite{var_pred}; spatiotemporal graph neural networks including DCRNN~\cite{li2018diffusion}, STGCN~\cite{yu2018spatio}, ASTGCN~\cite{guo2019attention}, GWNet~\cite{wu2019graph}, STGODE~\cite{fang2021spatial}, DSTAGNN~\cite{lan2022dstagnn}, ASTGNN~\cite{guo2021learning}, and AGCRN~\cite{bai2020adaptive}; spatially enhanced or embedding/normalization-based models including ST-WA~\cite{cirstea2022towards}, ST-Norm~\cite{deng2021st}, and STID~\cite{shao2022spatial}; transformer-based or attention-driven PDFormer~\cite{jiang2023pdformer} and STAEformer~\cite{liu2023spatio}; pre-training and masked-reconstruction STEP~\cite{STEP} and STD-MAE~\cite{STDMAE}; and heterogeneity or decomposition models HimNet~\cite{dong2024heterogeneity} and STDN~\cite{Cao_Wang_Jiang_Yu_Dong_2025}, ensuring comprehensive and fair assessment across modeling paradigms.

\subsubsection{Parameter Settings}
Following the BasicTS heterogeneity analysis protocol~\citep{shao2023exploring}, we set the similarity thresholds to $\epsilon_u=0.9$ and $\epsilon_l=0.5$. We also use a batch size of 8, 100 pre-training epochs, and at most 300 forecasting epochs. The specific hyperparameters are summarized in Table~\ref{hyper}.

\begin{table}[ht]
    \centering
    \caption{Detailed configurations of STOT and main baselines.}
    \small
    \renewcommand{\arraystretch}{1.05}
    \setlength{\tabcolsep}{3pt}
    \resizebox{\linewidth}{!}{%
    \begin{tabular}{lcccccc}
        \hline
        \textbf{Setting} & \textbf{STOT} & \textbf{STD-MAE} & \textbf{STEP} & \textbf{ST-WA} & \textbf{STID} & \textbf{ST-Norm} \\ \hline
        Learning Rate    & 0.001 & 0.001  & 0.001 & 0.001 & 0.002  & 0.002          \\
        Hidden Dimension    & 96 & 96  & 96  & 128 & 32  & \textendash      \\
        Number of Heads    & 4 & 4  & 4 & \textendash  & \textendash  & \textendash         \\
        Similarity Thresholds $(\epsilon_u,\epsilon_l)$    & (0.9, 0.5) & \textendash  & \textendash & \textendash  & \textendash  & \textendash         \\
        Mask Ratio    & 0.25 & 0.25  & 0.75 & \textendash  & \textendash  & \textendash   \\
        Sinkhorn Temperature $\mathbf{\epsilon}$    & 0.8 &  \textendash & \textendash & \textendash & \textendash  & \textendash   \\
        Sinkhorn Iteration Count K  & 3 & \textendash & \textendash & \textendash & \textendash & \textendash \\
        Batch Consistency Weight $\lambda$   & 0.7 & \textendash & \textendash & \textendash & \textendash & \textendash      \\ \hline
    \end{tabular}%
    }
    \label{hyper}
\end{table}

\begin{figure}[H]
    \centering
    \includegraphics[width=0.95\textwidth]{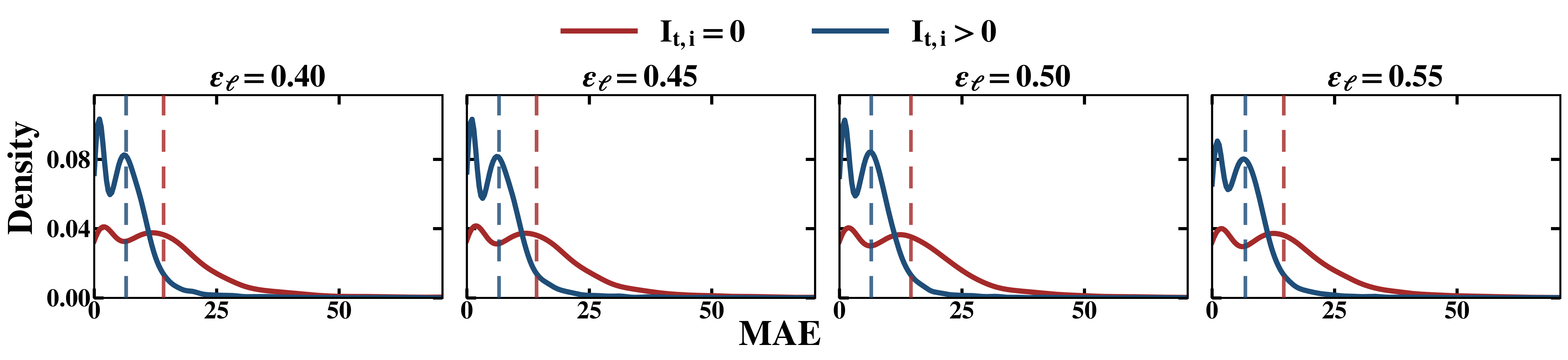}
    \caption{KDE distributions of node-sample forecasting MAE grouped by the spatial indistinguishability score in Equation 5.
    Dashed lines indicate the group-wise mean MAE.
    }
    \label{fig:formula5_kde}
\end{figure}

We further examined whether the low similarity threshold affects the validity of the spatial indistinguishability score. Because the score in Equation 5 is computed as a ratio between confidently dissimilar and confidently similar neighbors, its value depends on the overall separation of the similarity distribution and stays stable when the thresholds are perturbed within a reasonable range. To verify this property on a concrete model, we conducted a control analysis on the GWNet backbone~\cite{wu2019graph}. For each node sample, we computed $I_{t,i}$ and split the samples into one set with $I_{t,i}=0$ and one set with $I_{t,i}>0$, then compared their forecasting MAE distributions under $\epsilon_l \in \{0.40,0.45,0.50,0.55\}$. As shown in Figure~\ref{fig:formula5_kde}, the set with $I_{t,i}>0$ consistently attains larger MAE, which indicates that the score captures prediction difficulty and that the default threshold remains reliable.

\subsubsection{Evaluation Metrics}
We use three standard metrics ($y_i$: ground truth, $\hat{y}_i$: prediction): \textbf{MAE} $= \frac{1}{n}\sum|y_i - \hat{y}_i|$, \textbf{RMSE} $= \sqrt{\frac{1}{n}\sum(y_i-\hat{y}_i)^2}$, and \textbf{MAPE} $= \frac{1}{n}\sum\left|\frac{y_i-\hat{y}_i}{y_i}\right|\times 100$.
\subsubsection{Implementation Details}
We conducted experiments on 8 RTX A6000 GPUs and implemented our proposed model based on BasicTS~\citep{shao2023exploring}, which provides a unified framework for data loading and evaluation metrics to ensure fair comparisons. The shared training protocol and method-specific hyperparameters are detailed in the Parameter Settings subsection and summarized in Table~\ref{hyper}. For a fair comparison, all pre-training methods, including STEP, STD-MAE, and STOT, were run with the same pre-training length of $T_{long}=288$ steps, even though their original papers adopt longer sequences.

\subsection{Main Results (RQ1)}
\subsubsection{Comparison Results}
\begin{figure}[htbp]
    \centering
    \includegraphics[width=\linewidth]{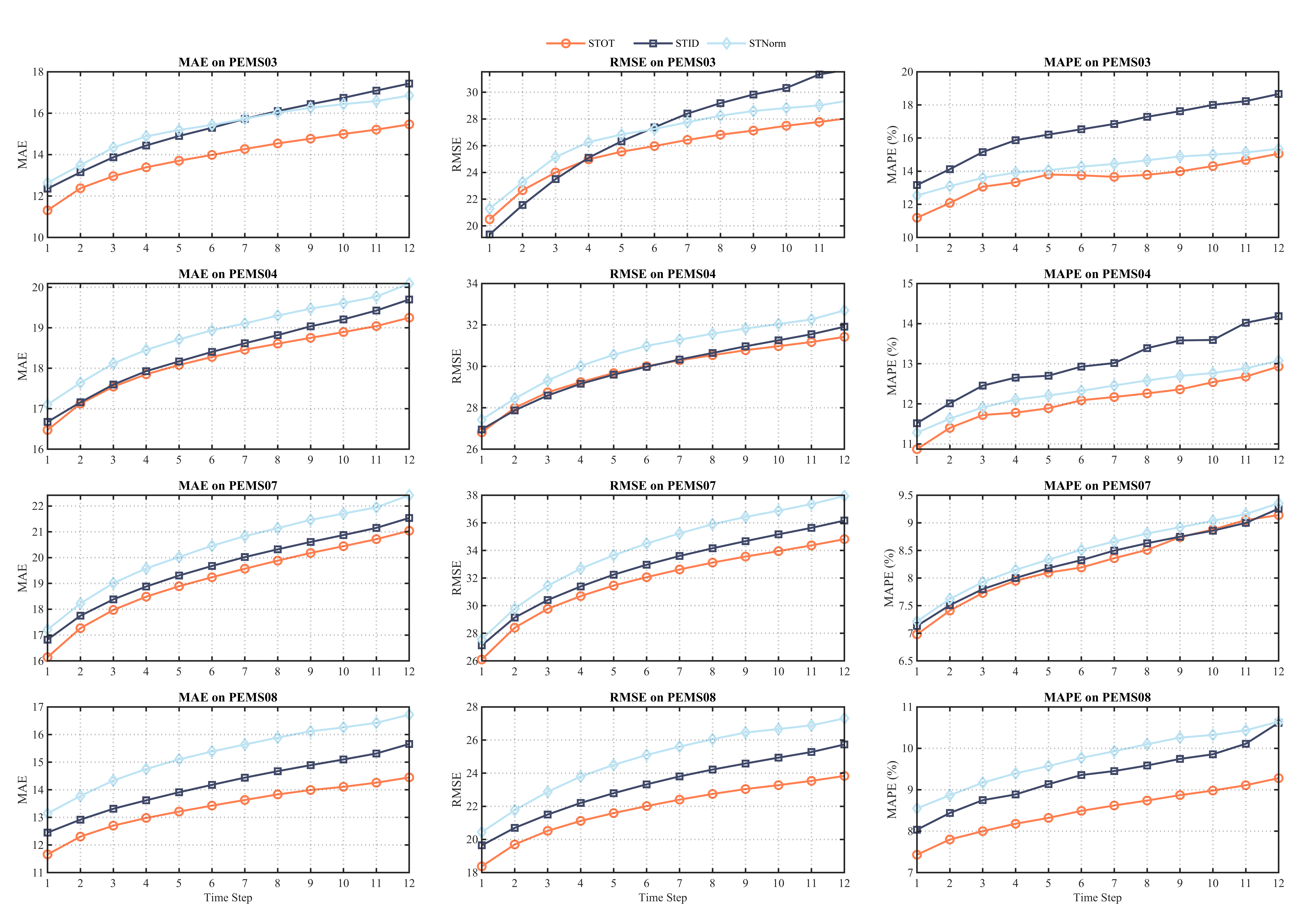}
    \caption{Multi-step prediction comparison on PeMS03, PeMS04, PeMS07, and PeMS08.}
    \label{fig:comp1}
\end{figure}

Tables \ref{compare1} and \ref{compare2} report six-dataset prediction results, with bold values marking the best performance; STOT consistently outperforms baselines on traffic speed and flow, and Figure~\ref{fig:comp1} shows that it surpasses STID and ST-Norm across 1- to 12-step forecasts on PeMS03/04/07/08, demonstrating its advantage in handling spatial indistinguishability. The observations are that conventional statistical methods struggle with nonlinear and non-stationary data, yielding lower accuracy than deep learning approaches; GNN-based methods such as AGCRN capture spatiotemporal correlations but suffer from RNN-based multi-step error accumulation, while attention-based PDFormer mitigates this issue yet still underperforms because of insufficient spatial modeling. Their message passing also aggregates signals over fixed neighborhoods, so the confusion carried by spatially indistinguishable nodes propagates to neighbors and reinforces ambiguous representations, which is the core difficulty that our masking strategy targets. Spatially-aware models such as STID and ST-Norm generalize better by directly addressing spatial distinguishability, outperforming GNN-based methods in accuracy and efficiency, while pre-trained STEP and STD-MAE learn from long historical sequences and focus on hard-to-predict nodes, with STD-MAE's dual autoencoders further exploiting spatial structure to achieve near-optimal results.
STOT's gains come from long-term inputs with temporal periodic features for richer spatiotemporal representations, adaptive spatial representations learned through mutual reconstruction guided by spatial distinguishability, and OT-based batch consistency that improves semantic coherence and generalization. The gain is smaller on PeMS-BAY because its nodes share very similar long-term levels and form few strongly correlated pairs, so the dataset carries less spatial indistinguishability for the optimal transport masking to exploit.

\begin{table}[htbp]
\caption{Performance comparison on PeMS03, PeMS04, PeMS07, and PeMS08 benchmarks.}
	\renewcommand{\arraystretch}{0.7}
    \setlength{\tabcolsep}{4pt}
	\centering
	\resizebox{1.0\linewidth}{!}{
    \begin{threeparttable}
\begin{tabular}{lccclccclccc}
\cline{1-12}
\multirow{2}{*}{Model}                                        & \multicolumn{3}{c}{PeMS03}                                                  &  & \multicolumn{3}{c}{PeMS04}                                                  &  & \multicolumn{3}{c}{PeMS07}                                                 \\ \cline{2-4} \cline{6-8} \cline{10-12}
                                                              & MAE                     & RMSE                    & MAPE                    &  & MAE                     & RMSE                    & MAPE                    &  & MAE                     & RMSE                    & MAPE                   \\ \cline{1-12}
ARIMA~\cite{ARIMA_pred}            & 35.31                   & 47.59                   & 33.78                   &  & 33.73                   & 48.80                   & 24.18                   &  & 38.17                   & 59.27                   & 19.46                  \\
VAR~\cite{var_pred}              & 23.65                   & 38.26                   & 24.51                   &  & 23.75                   & 36.66                   & 18.09                   &  & 75.63                   & 115.24                  & 32.22                  \\
DCRNN~\cite{li2018diffusion}            & 18.18                   & 30.31                   & 18.91                   &  & 24.70                   & 38.12                   & 17.12                   &  & 25.30                   & 38.58                   & 11.66                  \\
STGCN~\cite{yu2018spatio}               & 17.49                   & 30.12                   & 17.15                   &  & 22.70                   & 35.55                   & 14.59                   &  & 25.38                   & 38.78                   & 11.08                  \\
ASTGCN~\cite{guo2019attention}          & 17.69                   & 29.66                   & 19.40                   &  & 22.93                   & 35.22                   & 16.56                   &  & 28.05                   & 42.57                   & 13.92                  \\
GWNet~\cite{wu2019graph}                & 19.85                   & 32.94                   & 19.31                   &  & 25.45                   & 39.70                   & 17.29                   &  & 26.85                   & 42.78                   & 12.12                  \\
STGODE~\cite{fang2021spatial}           & 16.50                   & 27.84                   & 16.69                   &  & 20.84                   & 32.82                   & 13.77                   &  & 22.99                   & 37.54                   & 10.14                  \\
DSTAGNN~\cite{lan2022dstagnn}           & 15.57                   & 27.21                   & 14.68                   &  & 19.30                   & 31.46                   & 12.70                   &  & 21.42                   & 34.51                   & 9.01                   \\
ST-WA~\cite{cirstea2022towards}         & 15.17                   & 26.63                   & 15.83                   &  & 19.06                   & 31.02                   & 12.52                   &  & 20.74                   & 34.05                   & 8.77                   \\
ASTGNN~\cite{guo2021learning}           & 15.07                   & 26.88                   & 15.80                   &  & 19.26                   & 31.16                   & 12.65                   &  & 22.23                   & 35.95                   & 9.25                   \\
AGCRN~\cite{bai2020adaptive}            & 16.06                   & 28.49                   & 15.85                   &  & 19.83                   & 32.26                   & 12.97                   &  & 21.29                   & 35.12                   & 8.97                   \\
ST-Norm ~\cite{deng2021st}               & 15.32                   & 25.93                   & 14.37                   &  & 19.21                   & 32.30                   & 13.05                   &  & 20.59                   & 34.86                   & 8.61                   \\
STID ~\cite{shao2022spatial}                       & 15.30                   & 27.26                   & 16.47                   &  & 18.40                   & 29.98                   & 12.93                   && 19.61  &      32.83                   &    8.33                           \\
STEP * ~\cite{STEP}                 & 14.22                   & 24.55                   & 14.42                   &  & 18.20                   & 29.71                   & 12.48                   &  & 19.32                   & 32.19                   & 8.12                   \\
PDFormer~\cite{jiang2023pdformer}       & 14.94                   & 25.39                   & 15.82                   &  & 18.32                   & 29.97                   & 12.10                   &  & 19.83                   & 32.87                   & 8.53                   \\
STAEformer~\cite{liu2023spatio}         & 15.35                   & 27.55                   & 15.18                   &  & 18.22                   & 30.18                   & 11.98                   &  & 19.14                   & 32.60                   & 8.01                   \\
 HimNet \cite{dong2024heterogeneity}   &   15.34&27.50&N/A&&18.24&30.12&N/A&&19.30&32.77&N/A
 \\
STDN \cite{Cao_Wang_Jiang_Yu_Dong_2025}   &     15.54&27.28&N/A&&18.40&30.22&N/A&&20.65&34.77&N/A
 \\
STD-MAE* ~\cite{STDMAE}                  & 14.01                   & 25.79                   & 14.30                   &  & 18.10                   & 29.68                   & 11.97                   &  & 19.25 & 32.27 & 8.04 \\

\cline{1-12}
STOT (Our Study)                                                  & \bf{13.91} & \bf{25.71} & \bf{13.56} &  & \bf{18.08} & \bf{29.67} & \bf{11.89} &  & \bf{18.95} & \bf{31.63} & \bf{7.98} \\ \cline{1-12}
\end{tabular}


\end{threeparttable}}
\label{compare1}
\end{table}
\begin{table}[htbp]
\caption{Performance comparison on PeMS08, METR-LA, and PeMS-BAY benchmarks.}
	\renewcommand{\arraystretch}{0.7}
    \setlength{\tabcolsep}{4pt}
	\centering
	\resizebox{1.0\linewidth}{!}{
    \begin{threeparttable}
\begin{tabular}{llllllllllll}
\hline
\multirow{2}{*}{Model}                                        & \multicolumn{3}{c}{PeMS08}                                                    & \multicolumn{1}{c}{} & \multicolumn{3}{c}{METR-LA}                                                   &  & \multicolumn{3}{c}{PeMS-BAY}                                                  \\ \cline{2-4} \cline{6-8} \cline{10-12}
                                                              & \multicolumn{1}{c}{MAE} & \multicolumn{1}{c}{RMSE} & \multicolumn{1}{c}{MAPE} &                      & \multicolumn{1}{c}{MAE} & \multicolumn{1}{c}{RMSE} & \multicolumn{1}{c}{MAPE} &  & \multicolumn{1}{c}{MAE} & \multicolumn{1}{c}{RMSE} & \multicolumn{1}{c}{MAPE} \\ \hline
ARIMA~\cite{ARIMA_pred}            & 31.09                   & 44.32                    & 22.73                    &                      & 5.15                    & 10.45                    & 12.70                    &  & 2.33                    & 4.76                     & 5.40                     \\
VAR~\cite{var_pred}              & 23.46                   & 36.33                    & 15.42                    &                      & 5.28                    & 9.06                     & 12.50                    &  & 2.24                    & 4.96                   & 4.83                     \\
DCRNN~\cite{li2018diffusion}            & 17.86                   & 27.83                    & 11.45                    &                      & 3.59                    & 7.61                     & 10.44                    &  & 1.96                    & 4.59                     & 4.68                     \\
STGCN~\cite{yu2018spatio}               & 18.02                   & 27.83                    & 11.40                    &                      & 3.60                    & 7.50                     & 10.56                    &  & 1.99                    & 4.51                     & 4.66                     \\
ASTGCN~\cite{guo2019attention}          & 18.61                   & 28.16                    & 13.08                    &                      & 3.57                    & 7.19                     & 10.32                    &  & 1.86                    & 4.07                     & 4.27                     \\
GWNet~\cite{wu2019graph}                & 19.13                   & 31.05                    & 12.68                    &                      & 4.90                    & 9.70                     & 14.75                    &  & 2.71                    & 6.25                     & 6.69                     \\
STGODE~\cite{fang2021spatial}           & 16.81                   & 25.97                    & 10.62                    &                      & 4.73                    & 7.60                     & 11.71                    &  & 1.77                    & 3.89                     & 4.02                     \\
DSTAGNN~\cite{lan2022dstagnn}           & 15.67                   & 24.77                    & 9.94                     &                      & 3.32                    & 6.68                     & 9.31                     &  & 1.89                    & 4.11                     & 4.26                     \\
ST-WA~\cite{cirstea2022towards}         & 15.41                   & 24.62                    & 9.94                     &                      & 3.65                    & 7.56                     & 10.35                    &  & 2.01                    & 4.57                     & 4.70                     \\
ASTGNN~\cite{guo2021learning}           & 15.98                   & 25.67                    & 9.97                     &                      & N/A                     & N/A                      & N/A                      &  & N/A                     & N/A                      & N/A                      \\
AGCRN~\cite{bai2020adaptive}            & 15.95                   & 25.22                    & 10.09                    &                      & 3.68                    & 7.74                     & 10.82                    &  & 2.01                    & 4.64                     & 4.71                     \\
ST-Norm ~\cite{deng2021st}               & 15.39                   & 24.80                    & 9.91                     &                      & 3.56                    & 7.48                     & 10.20                    &  & 1.92                    & 4.49                     & 4.54                     \\
STID ~\cite{shao2022spatial}                       & 14.20                   & 23.29                    & 9.33                     &                      & 3.55                    & 7.55                     & 10.94                    &  & 1.90                    & 4.41                     & 4.44                     \\
STEP*~\cite{STEP}                 & 14.00                   & 23.41                    & 9.50                     &                      & 3.37                    & 6.99                     & 9.61                     &  & \bf{1.79}                    & 4.20                     & 4.18                     \\
PDFormer~\cite{jiang2023pdformer}       & 13.58                   & 23.51                    & 9.05                     &                      & 3.62                    & 7.47                     & 10.91                    &  & 1.91                    & 4.43                     & 4.51                     \\
STAEformer~\cite{liu2023spatio}         & 13.46                   & 23.25                    & 8.88                     &                      & 3.34                    & 7.04                     & 9.78                     &  & 1.85                    & 4.31                     & 4.33                     \\
 HimNet \cite{dong2024heterogeneity}   &        13.56&23.36&N/A&& N/A&N/A&N/A&&N/A&N/A& N/A
 \\
STDN \cite{Cao_Wang_Jiang_Yu_Dong_2025}   &  14.25&24.39&N/A&&N/A &N/A &N/A&&N/A &N/A &N/A
 \\
STD-MAE* ~\cite{STDMAE}                  & 13.45                   & 22.47                    & 8.78                     &                      & 3.42                    & 7.08                     & 9.62                     &  & \bf{1.79}                    & 4.21                    & 4.17                     \\

 \hline
STOT (Our Study)                                                  & \bf{13.37}                   & \bf{21.90}                    & \bf{8.49}                     &  & \bf{3.27}         &             \bf{6.75}                        &\bf{9.15}               &                          & 1.83  &   \bf{4.20}                      &      \bf{4.16}                                        \\ \hline
\end{tabular}
\begin{tablenotes}
        \footnotesize
        \item * Due to the use of fewer pre-training steps, these metrics may differ from those reported in their paper.
\end{tablenotes}
\end{threeparttable}
}
\label{compare2}
\end{table}

\subsubsection{Reconstruction Results and Visualization}
\begin{figure}[ht]
    \centering
    \includegraphics[width=0.9\linewidth]{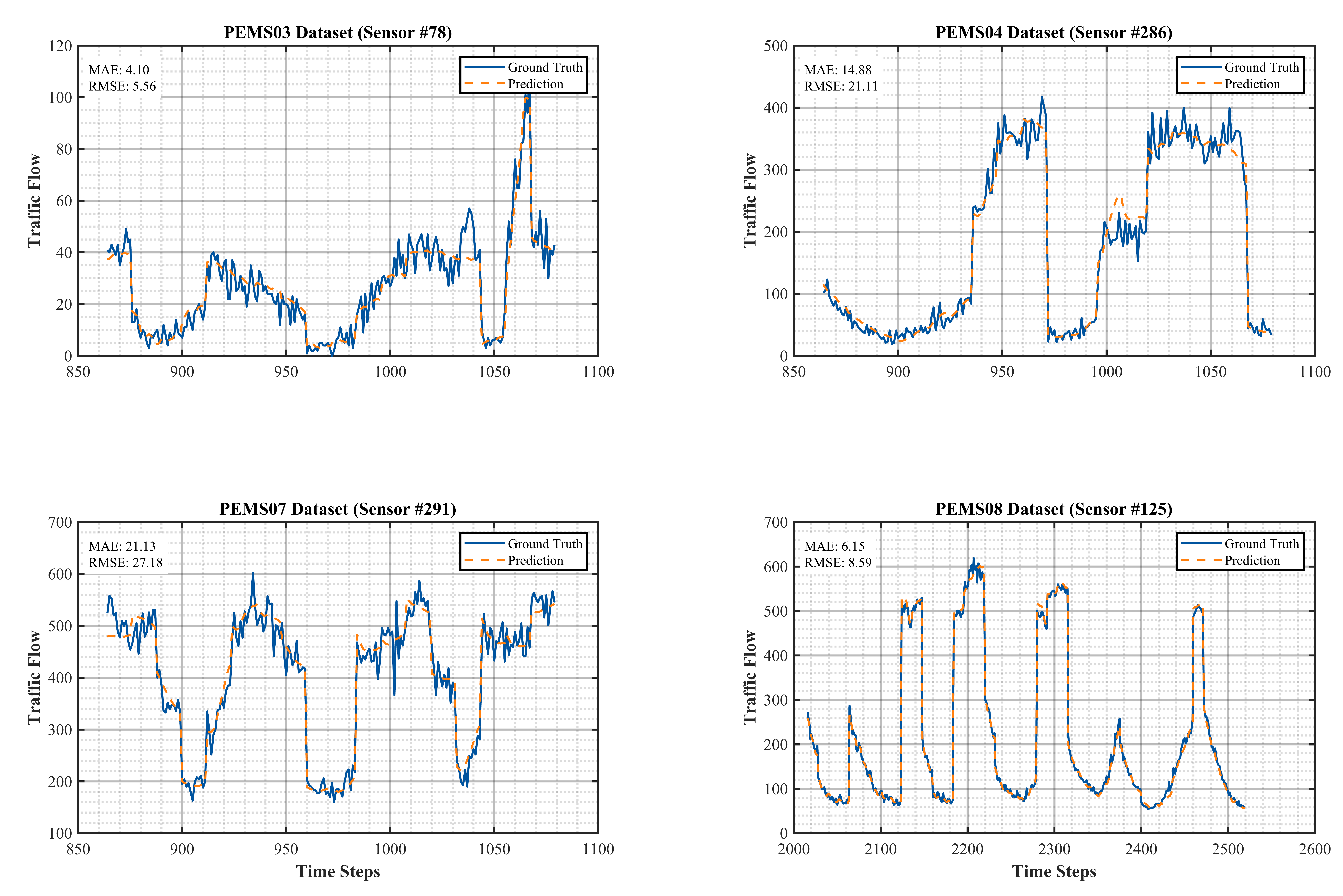}
    \caption{STOT reconstruction visualization on four datasets during pre-training.}
    \label{fig:res}
\end{figure}

Figure~\ref{fig:res} visualizes reconstruction accuracy on a randomly selected PeMS node, where STOT closely tracks the ground truth even under abrupt changes, such as the significant PeMS03 fluctuation during $t \in [850,1100]$, because its adaptive masking emphasizes spatially indistinguishable, high-variation timesteps.
\subsubsection{Forecasting Results and Visualization}
\begin{figure}[ht]
    \centering
    \includegraphics[width=0.9\linewidth]{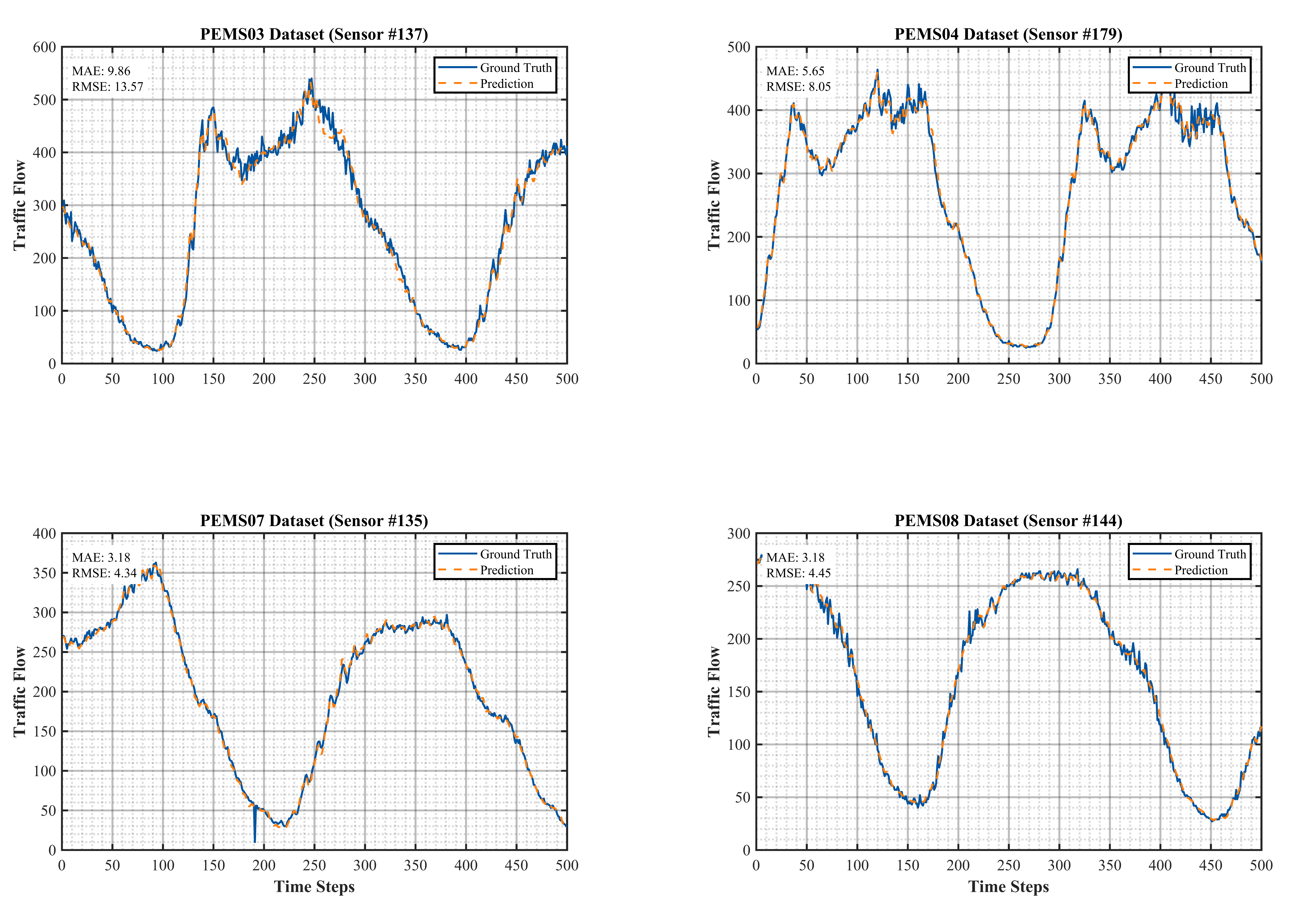}
    \caption{STOT forecasting visualization on four datasets after pre-training.}
    \label{fig:pre}
\end{figure}

Figure~\ref{fig:pre} shows forecasting accuracy on a randomly selected node, where predicted and true values align closely; the gain over the baseline without pre-training confirms that pre-training focuses the model on challenging nodes and improves its capture of complex spatiotemporal patterns.

\subsection{Ablation Study (RQ2)}
This section first evaluates the mask pre-training strategy in Section~\ref{mask}, then assesses STOT's generalization ability in Section~\ref{predictor}.

\subsubsection{Mask Ablation}
\label{mask}
To investigate each component of the designed mask pre-training strategy, we define five variants: \textbf{w/o Random walk mask} removes the post-OT random walk step and reduces mask-generation randomness; \textbf{w/o Optimal transport constraints} directly uses the most spatially indistinguishable positions as masks; \textbf{w/o Batch consistency constraints} allows masks at any position without penalty; \textbf{w/o Spatial indistinguishability guidance} removes this guidance and generates fully random masks; and \textbf{w/o Mask} applies no masked pre-training.

\begin{figure}[ht]
    \centering
    \includegraphics[width=0.9\linewidth]{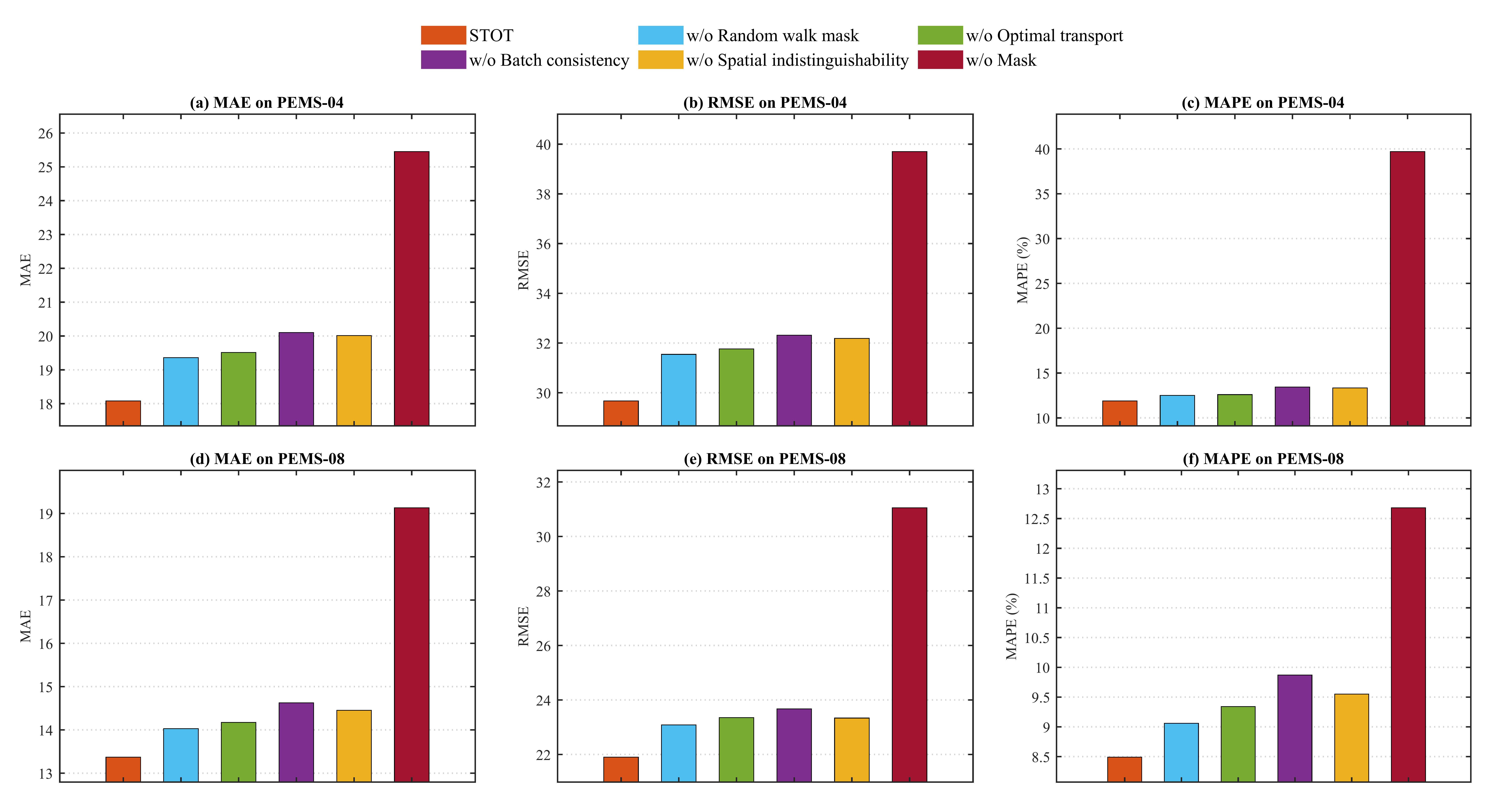}
    \caption{Mask pre-training component ablation on PeMS04 and PeMS08.}
    \label{fig:abs1}
\end{figure}

\begin{figure}[ht]
    \centering
    \includegraphics[width=0.9\linewidth]{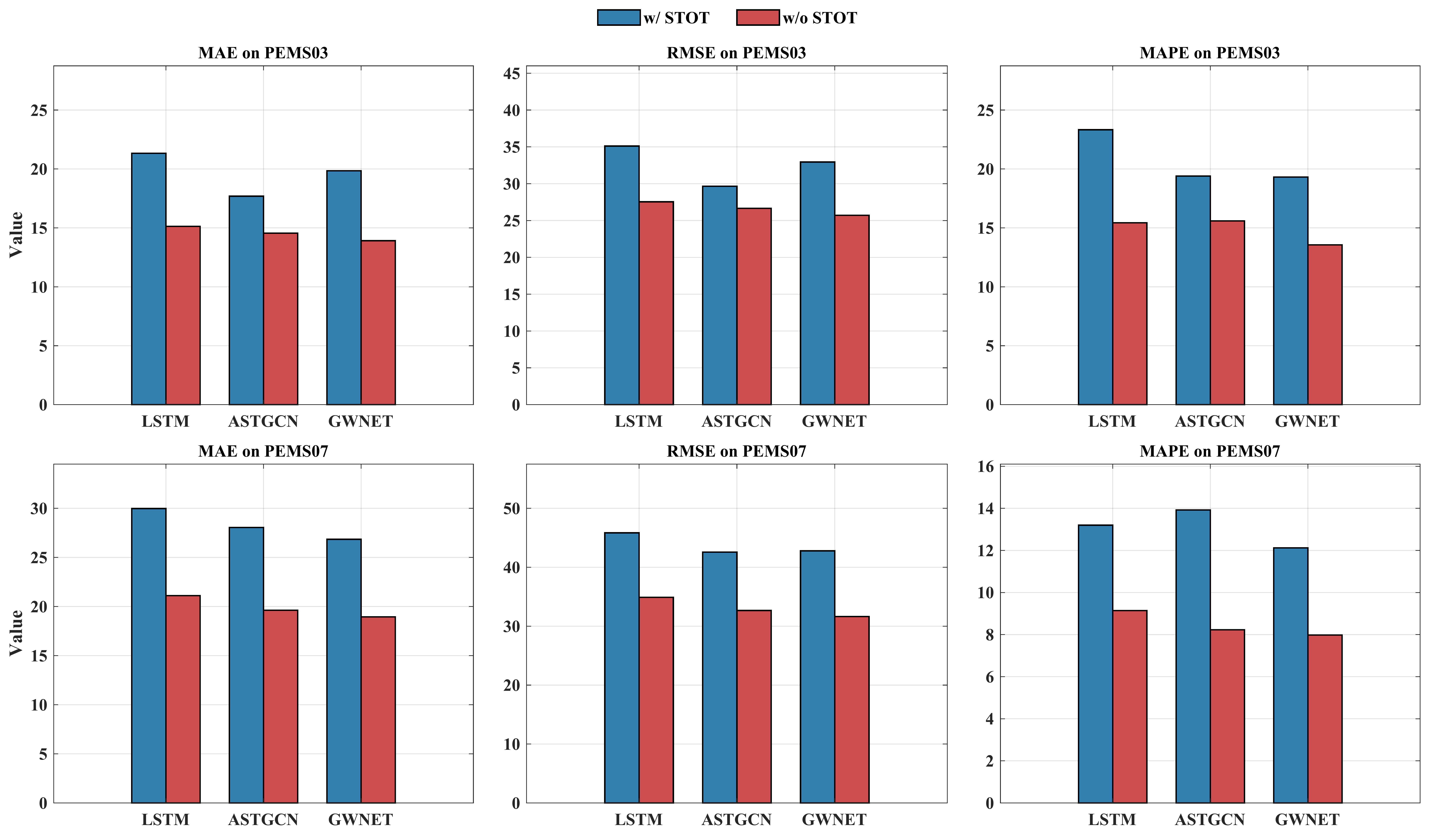}
    \caption{Mask pre-training component ablation on PeMS03 and PeMS07.}
    \label{fig:abs2}
\end{figure}
The ablation results in Figure~\ref{fig:abs1} show that removing the final random walk reduces mask diversity and creates static masks for identical sequences, hindering spatial dependency capture and confirming its role in spatial understanding; OT-based softening with a temperature parameter enables simultaneous temporal and spatial learning, replacing separate pre-training stages with an efficient single-step process; removing batch consistency causes severe degradation, which may indicate that semantic continuity is needed for coherent representations, accuracy, and robustness; eliminating spatial consistency supervision yields fully random masks and what may be trivial semantics, possibly reducing pre-training effectiveness and underscoring the need for guided mask generation; and removing masking strategies makes the model depend entirely on the predictor, highlighting the role of pre-training in accuracy, robustness, and generalization.
\subsubsection{Predictor Ablation}
\label{predictor}
To demonstrate that STOT enhances diverse backbones, we use it as a pre-trained feature extractor for LSTM, ASTGCN, and GWNet. The results in Figure~\ref{fig:abs2} show that STOT benefits downstream predictors despite architectural differences: before STOT, their performance differs substantially, whereas after integration the gap narrows, indicating stronger feature extraction, deeper spatiotemporal dependency modeling, and more accurate and robust prediction.

\subsection{High Spatial Indistinguishability Prediction Performance (RQ3)}
To evaluate performance under high spatial indistinguishability, we test STOT on PeMS04-s, a PeMS04 subset whose nodes have above-average spatial indistinguishability (Table~\ref{subdatasets}). Table~\ref{com2} summarizes the averaged results from three random experiments for all models, where STOT consistently outperforms baselines in longer-step prediction, captures long-term dependencies better, and maintains robust, stable performance under high indistinguishability, demonstrating reliable modeling of complex spatiotemporal dynamics when spatial patterns are difficult to discern.

\begin{table}[ht]
\centering
\caption{Dataset segmentation details.}
\renewcommand{\arraystretch}{1.2}
\setlength{\tabcolsep}{4pt}
\resizebox{\linewidth}{!}{
\begin{tabular}{lcccccc}
\hline
\textbf{Datasets} & \textbf{\#Sensors} & \textbf{\#Edges} & \textbf{\#Training} & \textbf{\#Validation} & \textbf{\#Testing} \\ \hline
PeMS04 (IN)    & 307 & 340 & [10173, 12, 307,1] & [3375, 12, 307,1]  & [3375, 12, 307,1]         \\
PeMS04 (OUT)    & 307 & 340 & [10173, 12, 307,1] & [3375, 12, 307,1]  & [3375, 12, 307,1]         \\
PeMS04-s (IN)    & 161 & 188 & [10173, 12, 161,1] & [3375, 12, 161,1]  & [3375, 12, 161,1]         \\
PeMS04-s (OUT)   & 161 & 188 & [10173, 12, 161,1] & [3375, 12, 161,1]  & [3375, 12, 161,1]         \\
\hline
\end{tabular}}
\label{subdatasets}
\end{table}

\begin{table}[ht]
    \centering
        \renewcommand{\arraystretch}{1.1}
        \setlength{\tabcolsep}{4pt}
    \caption{High spatial indistinguishability prediction on PeMS04.}
        \resizebox{0.90\linewidth}{!}{
    \begin{tabular}{llcccc}
    \hline
    \textbf{Model} & \textbf{Metrics} & \textbf{\#1 step} & \textbf{\#3 steps} & \textbf{\#6 steps} & \textbf{\#12 steps} \\
    \hline
    \multirow{3}{*}{STID} & MAE & 16.88±0.03 & 18.12±0.02 & 19.12±0.00 & 21.10±0.25 \\
    & RMSE & 27.29±0.05 & 29.31±0.03 & 30.82±0.01 & 33.32±0.27 \\
    & MAPE (\%) & 11.19±0.25 & 12.07±0.23 & 12.54±0.11 & 13.76±0.05 \\
    \hline
    \multirow{3}{*}{ST-Norm} & MAE & 16.26±0.02 & 18.23±0.03 & 19.58±0.07 & 21.66±0.14 \\
    & RMSE & 26.20±0.02 & 30.00±0.03 & 32.38±0.07 & 35.37±0.17 \\
    & MAPE (\%) & 7.18±0.21 & 8.01±0.23 & 8.41±0.27 & 9.55±0.32 \\
    \hline
    \multirow{3}{*}{STEP} & MAE & 16.88±0.02 & 18.13±0.01 & 19.13±0.00 & 20.88±0.10 \\
    & RMSE & 27.25±0.00 & 29.31±0.01 & 30.85±0.01 & 33.11±0.07 \\
    & MAPE (\%) & 11.66±0.16 & 12.46±0.10 & 12.97±0.08 & 14.50±0.35 \\
    \hline
    \multirow{3}{*}{STD-MAE} & MAE & 16.86±0.03 & 18.10±0.03 & 19.00±0.00 & 20.40±0.03 \\
    & RMSE & 27.30±0.03 & 29.27±0.03 & 30.68±0.04 & 32.62±0.11 \\
    & MAPE (\%) & 11.34±0.37 & 12.19±0.22 & 12.95±0.16 & 13.86±0.17 \\
    \hline
    \multirow{3}{*}{STOT} & MAE & 16.54±0.01 & 17.73±0.01 & 18.57±0.01 & 19.89±0.05 \\
    & RMSE & 26.85±0.00 & 28.87±0.00 & 30.23±0.03 & 31.97±0.11 \\
    & MAPE (\%) & 10.79±0.09 & 11.75±0.11 & 12.17±0.11 & 13.30±0.09 \\
    \hline
    \end{tabular}}
    \label{com2}
    \end{table}

\subsection{Hyperparameter Sensitivity (RQ4)}
\begin{figure}[htbp]
    \centering
    \includegraphics[width=1\linewidth]{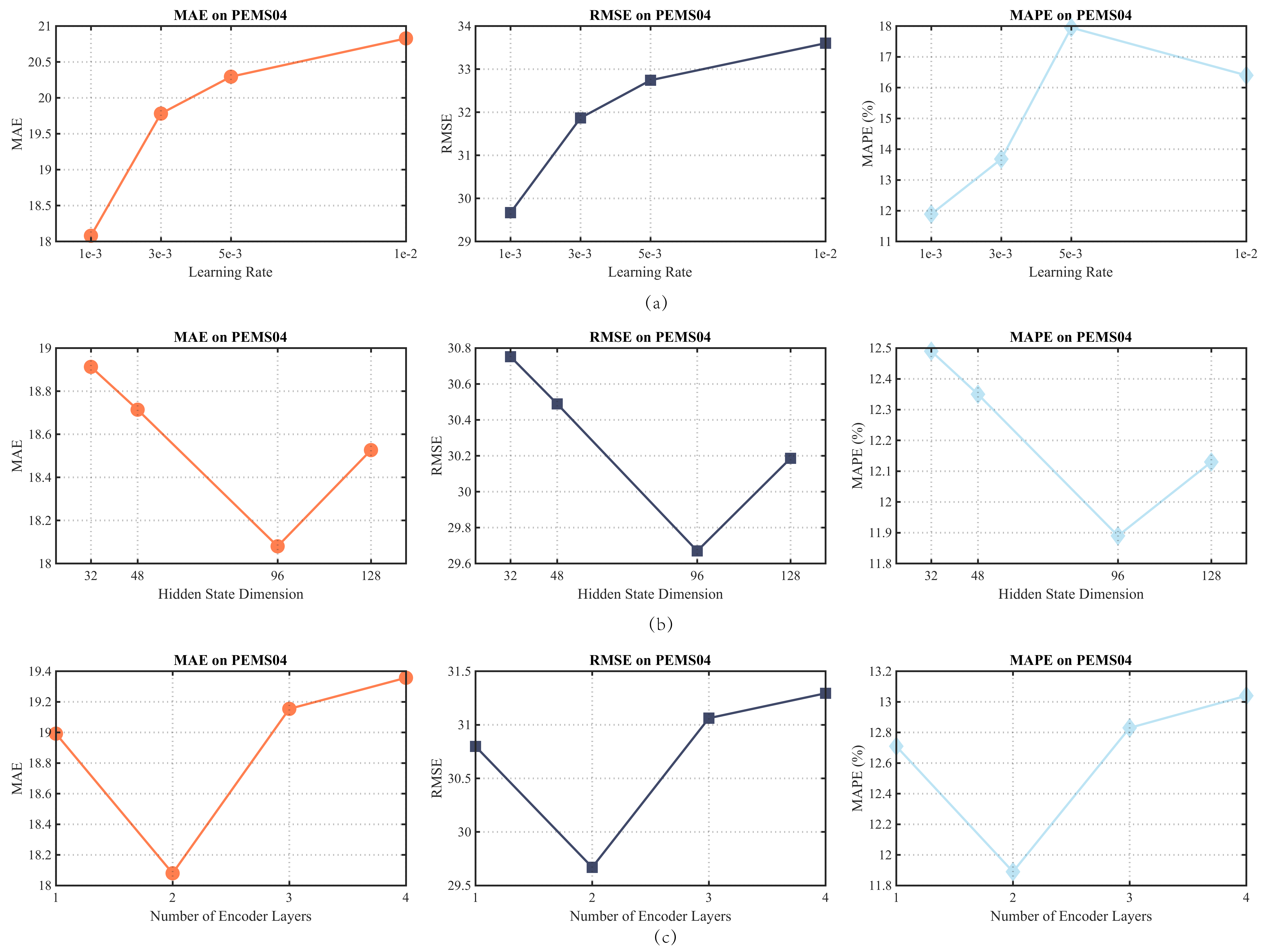}
    \caption{STOT parameter sensitivity on PeMS04 under MAE, RMSE, and MAPE.}
    \label{fig:hyper}
\end{figure}

We investigate STOT's sensitivity to learning rate, hidden dimension, and encoder depth. For \textbf{Learning Rate}, we test \{1\text{e}$^{-3}$, 3\text{e}$^{-3}$, 5\text{e}$^{-3}$, 1\text{e}$^{-2}$\}, and Figure~\ref{fig:hyper}(a) shows that accuracy decreases as the rate rises from 1\text{e}$^{-3}$ to 1\text{e}$^{-2}$, with 1\text{e}$^{-3}$ best balancing convergence speed and stability. For \textbf{Hidden Dimension}, we evaluate \{32, 48, 96, 128\}, and Figure~\ref{fig:hyper}(b) shows peak performance at 96, while larger dimensions slightly degrade results, suggesting that overly high-dimensional representations may introduce noise. For \textbf{Number of Encoder Layers}, we vary depth over \{1, 2, 3, 4\}, and Figure~\ref{fig:hyper}(c) shows that 2 layers perform best, indicating sufficient capacity for essential spatiotemporal dependencies while deeper encoders provide limited additional benefit.

\subsection{Computation Cost (RQ5)}
\begin{figure}[htbp]
    \centering
    \includegraphics[width=1\linewidth]{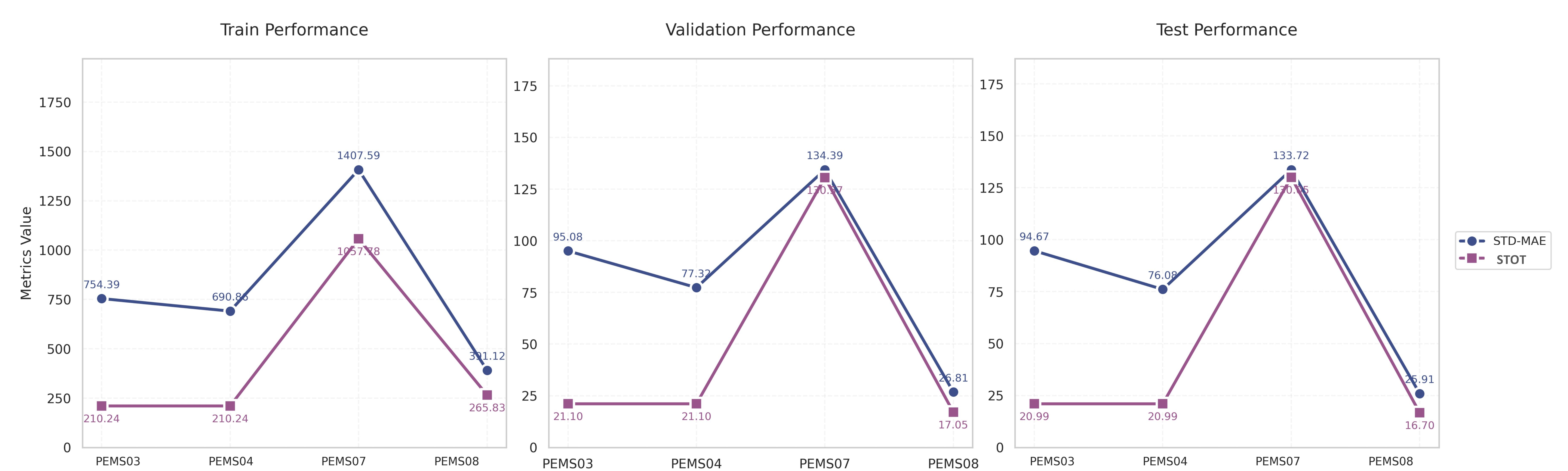}
    \caption{Time cost of different approaches.}
    \label{eff}
\end{figure}

Although STOT's autoencoder-based spatiotemporal representation learning may raise efficiency concerns, its dynamic mask strategy mitigates them. Figure~\ref{eff} reports per-sample pre-training and forecasting time on four datasets, showing that STOT is more efficient than other pre-training models, reduces spatial representation learning cost, and runs nearly twice as fast as STD-MAE while balancing competitive performance on the evaluated benchmarks and practical feasibility. For reproducibility, the timing comparison in Figure~\ref{eff} was measured on a single NVIDIA RTX A6000 GPU with a batch size of 8 for every model, matching the hardware reported in our experimental settings.

We further analyzed how mask generation scales with the number of sensor nodes, since the Sinkhorn solver is only one substep of the full mask generation pipeline. Table~\ref{tab:maskscale} reports, for each dataset, the node count together with the time of the complete mask generation and of the Sinkhorn substep alone, measured per batch on a single A6000 GPU. The full mask generation grows with the node count and reaches about 5.04 ms per batch on PeMS07 with 883 nodes, while the Sinkhorn substep stays close to 0.214 to 0.216 ms across all datasets. This stability arises because the optimal transport is solved on the temporal patch dimension after the node dimension has already been aggregated into the $Tscore$, so the solver cost is largely independent of the network size. These measurements indicate that the masking stays practical for large-scale urban sensor networks.

\begin{table}[htbp]
\centering
\caption{Mask generation time and the Sinkhorn substep time, in milliseconds per batch, across datasets with different numbers of nodes.}
\label{tab:maskscale}
\begin{tabular}{lccc}
\toprule
Dataset & Nodes & Mask generation & Sinkhorn \\
\midrule
PeMS08   & 170 & $1.3466 \pm 0.0852$ & $0.2138 \pm 0.0036$ \\
METR-LA  & 207 & $1.4644 \pm 0.1238$ & $0.2150 \pm 0.0035$ \\
PeMS04   & 307 & $1.3837 \pm 0.0891$ & $0.2137 \pm 0.0009$ \\
PeMS-BAY & 325 & $1.3890 \pm 0.0963$ & $0.2149 \pm 0.0034$ \\
PeMS03   & 358 & $1.4294 \pm 0.0556$ & $0.2154 \pm 0.0029$ \\
PeMS07   & 883 & $5.0419 \pm 0.3514$ & $0.2157 \pm 0.0038$ \\
\bottomrule
\end{tabular}
\end{table}

\subsection{OT Constraint Visualization (RQ6)}
\label{OT}
\begin{figure}[htbp]
    \centering
    \includegraphics[width=1\linewidth]{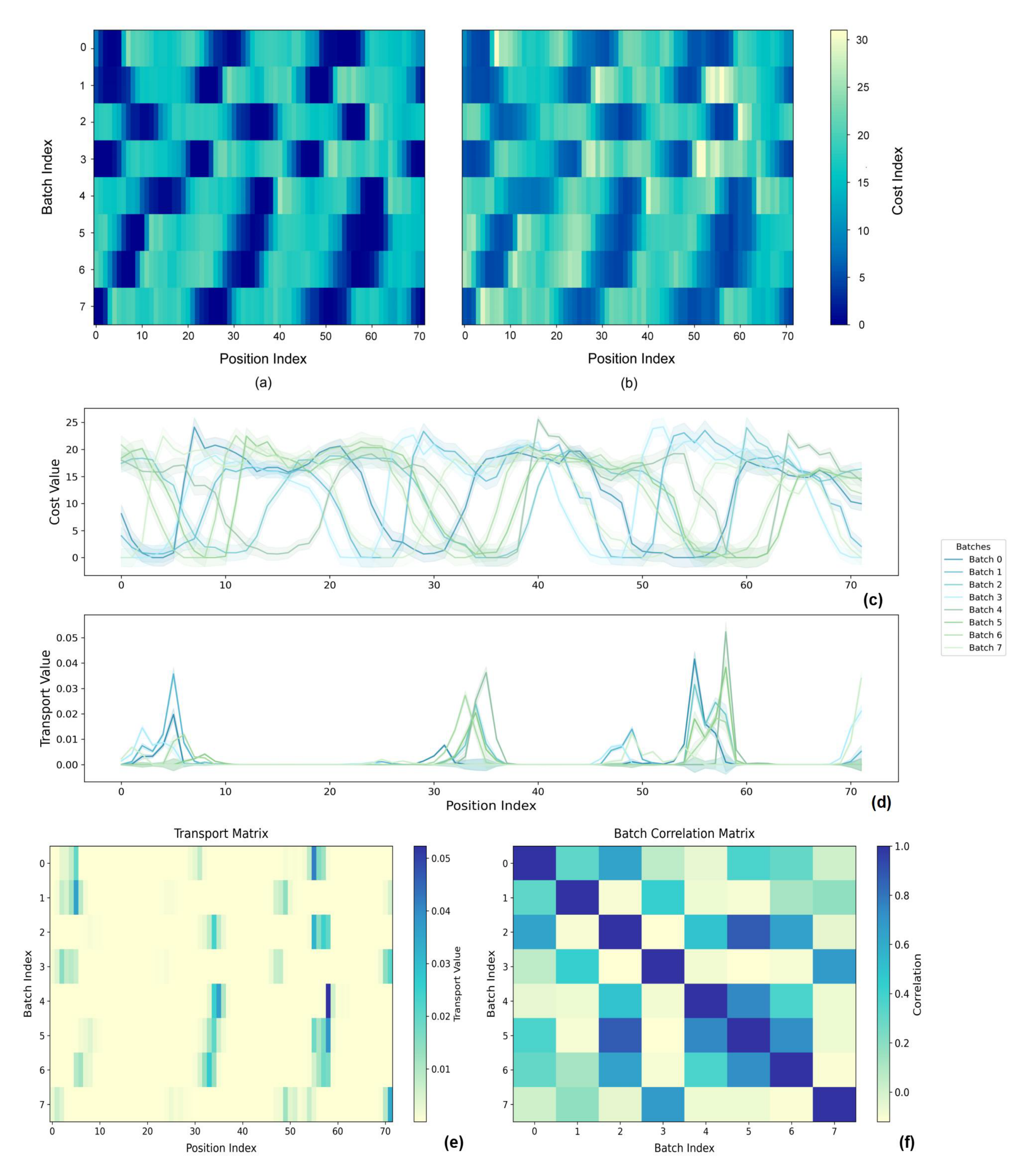}
    \caption{Visualization of OT and batch consistency: (a) and (b) cost matrices before and after batch consistency for the first 72 timesteps (6 patches), (c) and (d) the final cost matrix and OT matrix, (e) transport-matrix heatmap, and (f) batch-mask correlation matrix showing the effect of batch consistency on mask generation.}
    \label{fig:batch}
\end{figure}

To interpret STOT's OT effectiveness, we visualize the cost matrix, time-varying features, OT-generated masks, and batch consistency effects. Figure~\ref{fig:batch}(a) and Figure~\ref{fig:batch}(b) show the cost matrix before and after batch consistency constraints, with input timesteps on the horizontal axis, a batch size of 8 on the vertical axis, and color intensity indicating the cost magnitude that controls the likelihood of masking each position.

The original cost matrix in Figure~\ref{fig:batch}(a) has a highly concentrated color distribution; after applying batch consistency in Figure~\ref{fig:batch}(b), the distribution remains concentrated but gains more overlapping regions. These regions encourage masks at similar, though not identical, positions within a batch, which may enhance semantic understanding and improve downstream performance.

Figure~\ref{fig:batch}(c) shows that spatial indistinguishability has clear temporal periodicity: the lowest costs occur during morning and evening peaks, indicating the most severe indistinguishability and the greatest prediction challenge, while weaker off-peak fluctuations are less likely to be bottlenecks. This illustrates how SI-guided masking emphasizes challenging peak-hour segments, helping STOT learn richer spatiotemporal dependencies and improve accuracy.

Figure~\ref{fig:batch}(d) shows that transport-matrix peaks align with the low-cost regions in Figure~\ref{fig:batch}(c), validating the Sinkhorn algorithm and matching our design expectations. The peak positions also vary slightly across batches, with three distinct peaks in the time range [40, 60], which may reflect STOT's batch-adaptive mask adjustment for more accurate prediction.

The transport-matrix heatmap in Figure~\ref{fig:batch}(e) intuitively shows preferred mask locations, where masks are continuous within batches and periodic across timesteps. Figure~\ref{fig:batch}(f) further shows high inter-batch correlation among generated masks, which may facilitate optimization and enhance overall performance.

Together, these visualizations support the effectiveness of STOT's OT process: transport peaks align with low-cost regions, mask positions adjust dynamically across batches, masks remain continuous and periodic, and high inter-batch correlation improves optimization and prediction accuracy. These findings validate STOT's design choices, clarify its internal behavior, and motivate future improvements for spatiotemporal prediction.

\begin{figure}[htbp]
    \centering
    \includegraphics[width=1\linewidth]{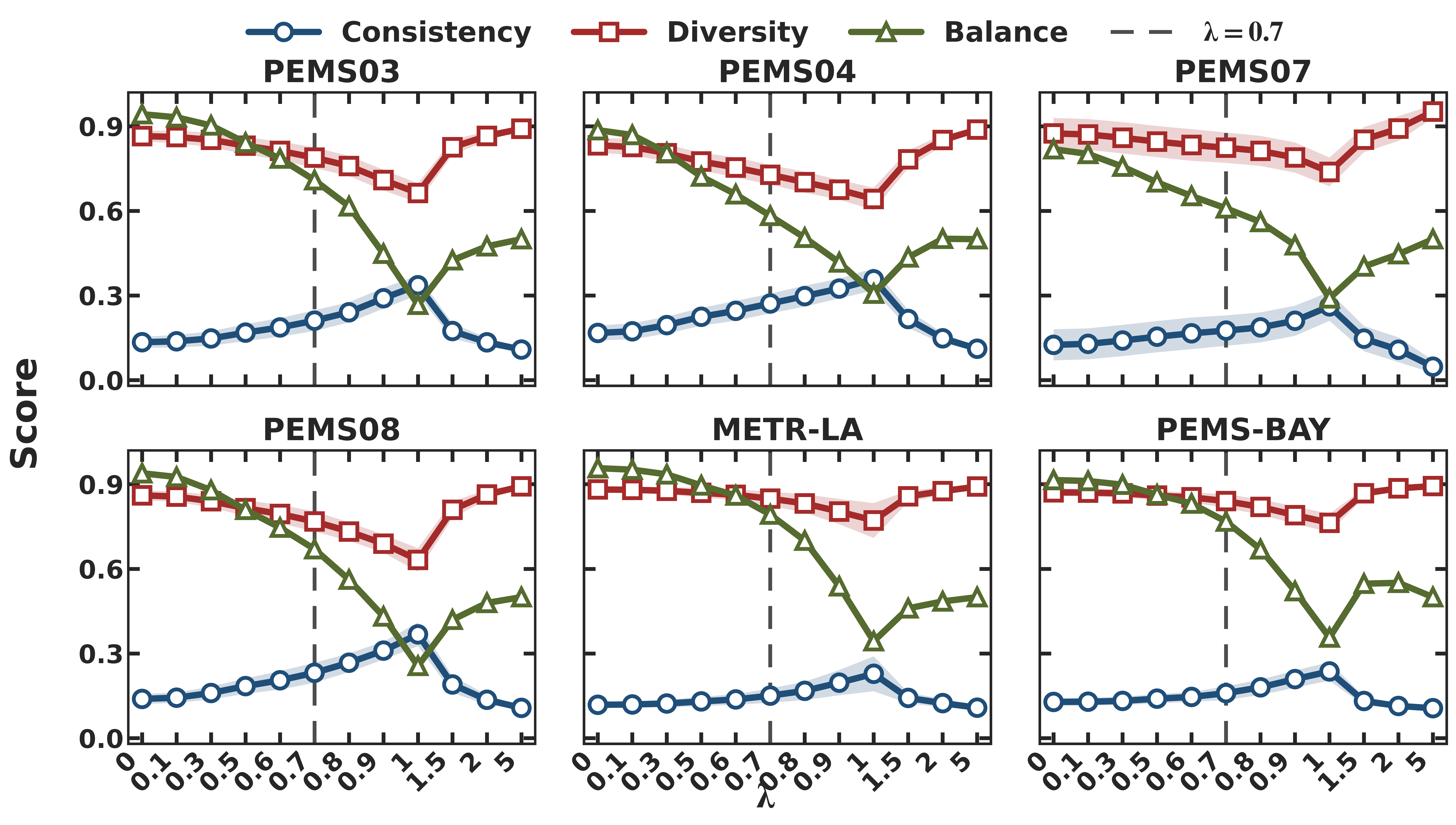}
    \caption{Consistency, diversity, and balance scores as functions of the batch consistency weight $\lambda$ on the six datasets. The dashed line marks the default $\lambda=0.7$.}
    \label{fig:lambda_tradeoff}
\end{figure}

We further quantified how the batch consistency weight $\lambda$ in Equation 7 trades off consistency against diversity. For a fixed training batch we regenerated the masks across a range of $\lambda$ without retraining, measured consistency as the average pairwise Jaccard similarity among the batch masks, used its complement as diversity, and combined the normalized diversity with the mean $Tscore$ of the selected patches into a balance score. As reported in Figure~\ref{fig:lambda_tradeoff}, the balance score stays near its peak while $\lambda$ remains small and declines once $\lambda$ exceeds 0.7, while consistency keeps rising and diversity keeps shrinking. The default $\lambda=0.7$ therefore sits at the point where the masks stay diverse enough to cover varied scenarios while the consistency term already aligns them across the batch, and this behavior holds on all six datasets.

\subsection{Random Walk Mechanism for Mask Generation (RQ7)}
\begin{figure}[htbp]
    \centering
    \includegraphics[width=1\linewidth]{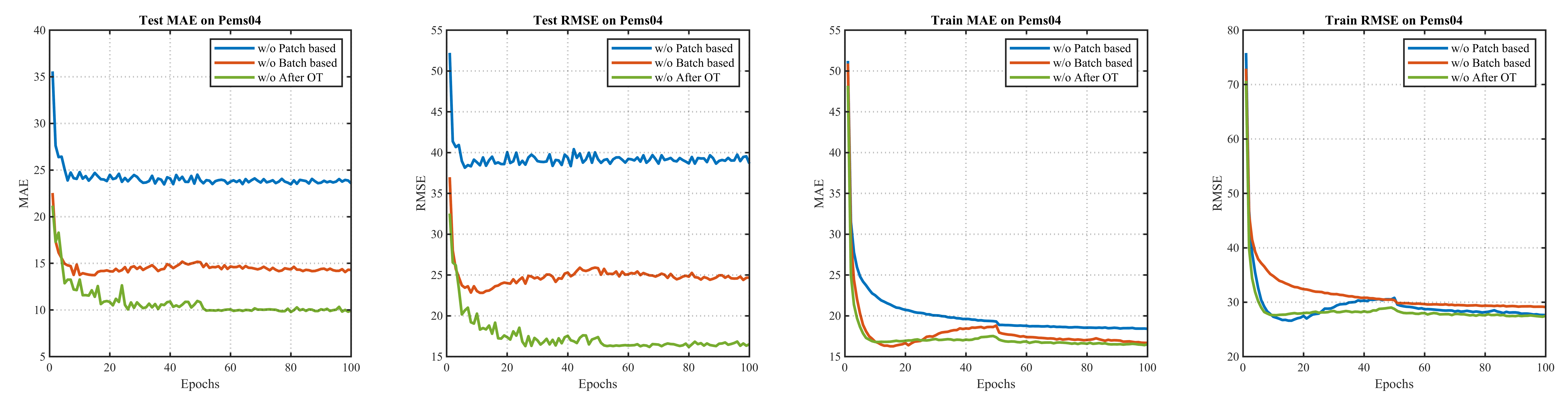}
    \caption{Comparison of random walk mechanism variants on PeMS04.}
    \label{fig:walk}
\end{figure}
To assess the random walk design, we test alternative implementations that operate on different positions or aspects and could substitute for the original mechanism:

\begin{itemize}
    \item \textbf{w/o Patch-based}: The smallest unit for the random walk is a single time step, as opposed to a patch of time steps.
    \item \textbf{w/o Batch-based}: The random walk does not apply simultaneously to the entire batch, but rather to individual samples within the batch.
    \item \textbf{w/o After OT}: The random walk is performed before the OT process, instead of after it.
\end{itemize}

Figure~\ref{fig:walk} reports training and testing losses for different random walk designs, leading to the following conclusions:

\begin{itemize}
    \item \textbf{w/o Patch-based}: Applying random walk at the timestep level simplifies inter-timestep semantics and makes deep representation learning harder; however, because batch consistency remains, optimization is not strongly affected, and the model resembles the T-MAE design in STD-MAE, achieving relatively good results among variants.

    \item \textbf{w/o Batch-based}: Disrupting batch consistency within one optimization step produces fragmented and possibly less meaningful semantics, so this design may underperform the Patch-based approach that preserves semantic continuity.

    \item \textbf{w/o After OT constraint}: Performing random walk before OT makes randomness depend on cost-matrix softening; temperature adjustment partly mitigates this, but mask-generation randomness is still greatly reduced, possibly leading to redundant semantics and the lowest performance.

\end{itemize}

Overall, all three variants converge to nearly identical training losses, which may indicate that they can almost fully reconstruct the sequence, while their performance differences arise from where random walk is applied. Our design avoids the above issues and enables comprehensive spatiotemporal modeling.

\section{Conclusion}
\label{con}
This paper introduced STOT, a self-supervised spatiotemporal prediction framework that improves representation learning under the common but underexplored challenge of spatial indistinguishability. Instead of treating masking as random or heuristic, STOT uses an optimal transport (OT)-based strategy to construct training signals aligned with intrinsic spatial structure. Across multiple real-world benchmarks, STOT improves over strong baselines, suggesting that explicitly modeling ambiguity among spatially similar regions may offer a forecasting advantage.

A key strength of our approach is that the OT-based masking mechanism provides both effectiveness and interpretability. Empirically, the strongest gains appear during peak hours, where demand patterns are volatile and prediction becomes most difficult, suggesting that the proposed masking strategy is especially beneficial when the model must discriminate among multiple plausible spatial explanations. The transport matrices offer an interpretable lens into how the model perceives "substitutable" or low-cost spatial regions, and the observed alignment between low-cost regions and high transport mass supports the intuition that OT can serve as a principled tool for constructing challenging but semantically meaningful self-supervised objectives. The batch consistency constraint further strengthens the learned representations by encouraging coherent semantics while allowing sufficient diversity, helping to stabilize training and improve generalization. Since the reconstruction objective draws on a longer context and a direct target, it is easier to optimize than forecasting, so the pre-training losses of different variants settle at close values, and the downstream gains instead come from masking that concentrates the reconstruction on the spatially indistinguishable nodes. However, our work has several limitations. The computational overhead introduced by OT can become a bottleneck when the number of spatial units grows or when extremely long sequences are used, raising scalability concerns for real-world deployments on city-scale or nation-scale grids. Additionally, STOT focuses primarily on endogenous spatiotemporal signals and may not fully capture the causal drivers of variation when exogenous variables (e.g., weather, holidays, accidents) dominate the dynamics. Finally, the current analysis does not fully characterize when OT-based masking is most beneficial, leaving open questions about robustness under distribution shifts. In addition, the optimal transport objective is easier to optimize under a moderate mask ratio, so STOT does not exploit the high mask ratios that benefit single-stage masking methods such as STEP~\cite{STEP}, which leaves room for transport formulations that stay stable under heavier masking.

Beyond performance, STOT offers two practical contributions: it provides a general framework for self-supervised objectives that address spatial ambiguity in traffic and environmental monitoring, with OT-based masking usable as a plug-in strategy under limited labels; and its transport matrices act as interpretable diagnostics for spatial similarity, data quality assessment, and decision policy design. Future work includes faster OT approximations for large-scale graphs, exogenous covariates for robustness, and extensions to anomaly detection, imputation, and cross-city transfer.
\section*{CRediT authorship contribution statement}
\textbf{Guangyu Wang:} Conceptualization, Methodology, Software, Validation, Formal analysis, Investigation, Data curation, Writing -- original draft, Visualization. \textbf{Jiawei Tong:} Conceptualization, Methodology, Writing -- review \& editing, Supervision, Project administration.

\section*{Acknowledgements}
The authors would like to thank Chen Yu for language polishing and proofreading, and Yujie Chen for assistance with manuscript formatting and typesetting.

\bibliographystyle{elsarticle-num-names}
\bibliography{main}

\end{document}